\documentclass{article}
\usepackage{iclr2024_conference,times}

\usepackage{amsmath,amsfonts,bm}

\def\eqref#1{equation~\ref{#1}}

\def\1{\bm{1}}

\DeclareMathAlphabet{\mathsfit}{\encodingdefault}{\sfdefault}{m}{sl}
\SetMathAlphabet{\mathsfit}{bold}{\encodingdefault}{\sfdefault}{bx}{n}

\usepackage{hyperref}
\usepackage{url}
\usepackage{gensymb}
\usepackage{amsmath}
\usepackage{multicol}
\usepackage{wrapfig}
\usepackage{array}
\usepackage{amsfonts,amssymb}
\usepackage{hyperref}
\usepackage{cleveref}
\usepackage{bm}
\usepackage{caption}
\usepackage{graphicx}
\usepackage{float}
\usepackage{booktabs}
\usepackage{multirow}
\usepackage{subfigure}
\usepackage{enumitem}
\usepackage[ruled,linesnumbered]{algorithm2e}
\crefname{equation}{Eq.}{Eqs.}
\crefname{figure}{Fig.}{Figs.}
\crefname{section}{Sec.}{Secs.}
\crefname{table}{Tab.}{Tabs.}
\usepackage{graphicx}
\usepackage[T1]{fontenc}

\usepackage{newfloat}
\usepackage{listings}

\title{InsertNeRF: Instilling Generalizability into NeRF with HyperNet Modules}

\author{Yanqi Bao$^{1}$, Tianyu Ding$^{2}$, Jing Huo$^{1}$\thanks{Corresponding author.}~, Wenbin Li$^1$, Yuxin Li$^{1}$, Yang Gao$^1$\\
{\normalsize{$^1$State Key Laboratory for Novel Software Technology, Nanjing University, Nanjing, China}}\\
\url{yq_bao@smail.nju.edu.cn, {huojing, liwenbin, gaoy} @nju.edu.cn,}\\
\url{liyuxin16@smail.nju.edu.cn}\\
{\normalsize{$^2$Applied Sciences Group, Microsoft Corporation, Redmond, USA}}\\
\url{tianyuding@microsoft.com}}

\begin{document}

\maketitle

\begin{abstract}

Generalizing Neural Radiance Fields (NeRF) to new scenes is a significant challenge that existing approaches struggle to address without extensive modifications to vanilla NeRF framework. We introduce \textbf{InsertNeRF}, a method for \textbf{INS}tilling g\textbf{E}ne\textbf{R}alizabili\textbf{T}y into \textbf{NeRF}. By utilizing multiple plug-and-play HyperNet modules, InsertNeRF dynamically tailors NeRF's weights to specific reference scenes, transforming multi-scale sampling-aware features into scene-specific representations. This novel design allows for more accurate and efficient representations of complex appearances and geometries. Experiments show that this method not only achieves superior generalization performance but also provides a flexible  pathway for integration with other NeRF-like systems, even in sparse input settings. 
Code will be available at: \url{https://github.com/bbbbby-99/InsertNeRF}.
\end{abstract}

\section{Introduction}

Novel view synthesis, a fundamental discipline in computer vision and graphics, aspires to create photorealistic images from reference inputs. Early works~\citep{debevec1996modeling, lin2004geometric} primarily focused on developing explicit representations, facing challenges due to the absence of 3D supervision. This issue has been alleviated by recent advancements in implicit neural representation research, which have led to improved performance.  In particular, Neural Radiance Fields~(NeRF)~\citep{mildenhall2021nerf} has attracted significant interest. NeRF, and its derivative works, extract scene-specific implicit representations through overfitting training on posed scene images. Although NeRF uses neural scene representations effectively to yield realistic images, the scene-specific nature of these representations requires retraining when faced with novel scenarios. 

{An emerging topic} known as Generalizable NeRF~(GNeRF) has recently garnered considerable attention {for this challenge.} GNeRF aims to learn a scene-independent inference approach that facilitates the transition from {references} to {target view}. Current methods enhance the NeRF architecture by adding structures that {aggregate reference-image features, or reference features}. Examples include {pixel-wise} feature cost volumes~\citep{johari2022geonerf}, transformers~\citep{wang2022attention,suhail2022generalizable}, and 3D visibility predictors~\citep{liu2022neural}. However, {fitting these additions into} conventional NeRF-like frameworks such as mip-NeRF~\citep{barron2021mip}, NeRF++~\cite{zhang2020nerf++}, and others, often proves challenging and may fail to effectively harness the guiding potential of reference features. Furthermore, the extensive use of transformers or cost volumes can be time-consuming. Thus, an intriguing question arises: Is it possible to directly \textbf{INS}till g\textbf{E}ne\textbf{R}alizabili\textbf{T}y into \textbf{NeRF} (\textbf{InsertNeRF}) while staying faithful to the original framework?

A straightforward way to accomplish this goal is to adaptively modify the NeRF network's weights, {or implicit representations}, for different reference scenes while preserving the original framework. The concept of hypernetwork~\citep{ha2016hypernetworks}, which conditionally parameterizes a target network, is an effective strategy in this scenario. The features extracted from the reference scene can be used as inputs to generate scene-specific network weights. However, initial experiments {in~\Cref{tab2}} indicate that constructing a hypernetwork directly based on the NeRF framework can be inadequate, and often fails to predict different attributes like emitted color and volume density. To address this, we propose to use \emph{HyperNet modules}, which are designed to serve as easily integrable additions to existing NeRF-like frameworks. 
Owing to their flexibility, the resulting InsertNeRF excels at predicting the NeRF attributes by capitalizing on sampling-aware features and various module structures.

\begin{figure}[t]
\vspace{-.4in}
\begin{center}
\includegraphics[width=1\textwidth]{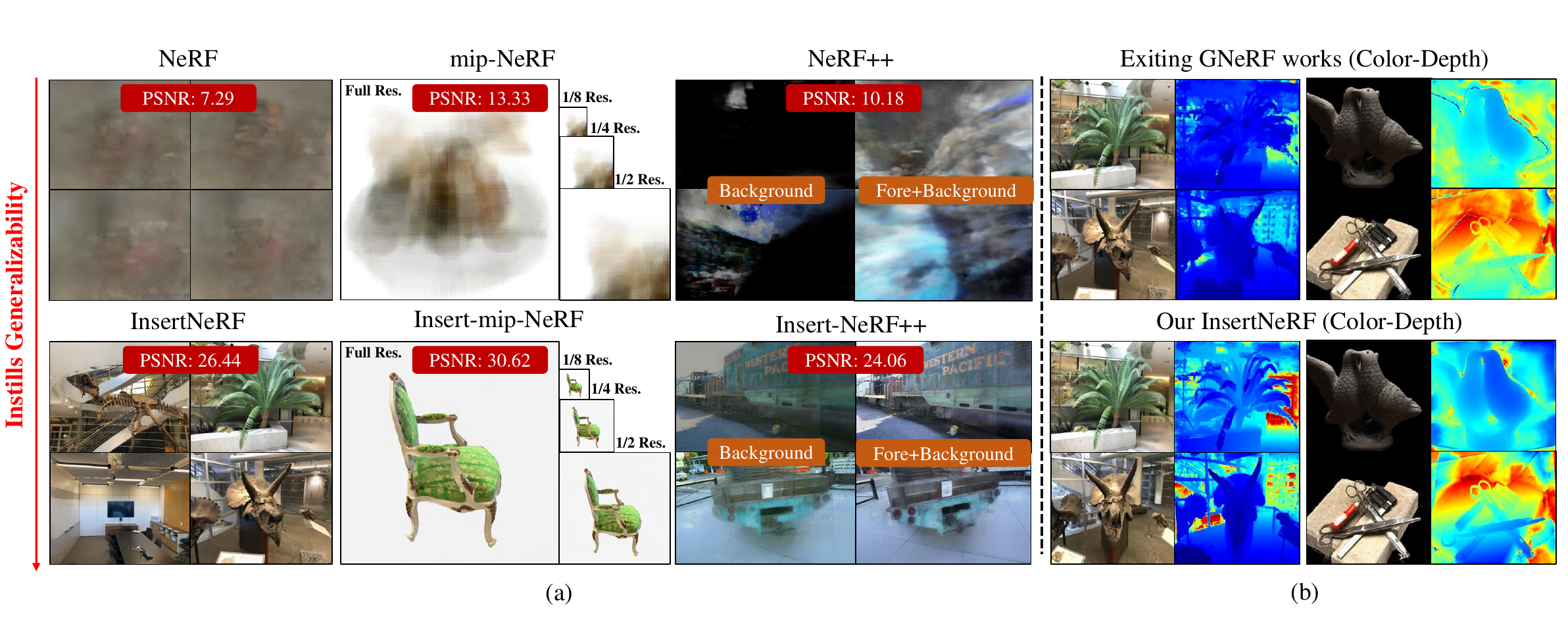}
\end{center}
\vspace{-.2in}
\caption{
{Overview of motivation.} (a) We instill generalizability into NeRF-like systems, including vanilla NeRF, mip-NeRF, {and NeRF++ frameworks}, to achieve consistent performance across scenes without modifying the base framework or requiring scene-specific retraining. (b) InsertNeRF significantly improves depth estimation compared to its original counterpart.
}
\label{fig1}
\vspace{-.2in}
\end{figure}

In InsertNeRF, we insert multiple HyperNet modules to instill generalizability throughout the framework's progression. This approach allows us to fully leverage the guiding role of scene features in determining the entire network's weights. Unlike existing works that solely utilize reference features as inputs, InsertNeRF exhibits a thorough grasp of \textit{reference scene knowledge}.
To further unlock the full potential of the HyperNet modules, it is crucial to aggregate scene features from a set of nearby reference images. To achieve this, we introduce a \emph{multi-layer dynamic-static} aggregation strategy. Compared to existing works, it not only harnesses the inherent completion capabilities of global features, but it also implicitly {models} occlusion through dynamic-static weights, as demonstrated on the depth renderings shown in~\Cref{fig1}b. By feeding the aggregated scene features into the HyperNet modules, we can generate scene-related weights based on the well-understood reference scene.


In summary, we make the following specific contributions:
\vspace{-.1in}
\begin{itemize}[leftmargin=.5cm]
    \item We introduce InsertNeRF, a novel paradigm that inserts multiple plug-and-play HyperNet modules into the NeRF-like framework, endowing NeRF-like systems with instilled generalizability.
    \item We design two types of HyperNet module structures tailored to  different NeRF attributes, aiming for predicting scene-specific weights derived from sampling-aware scene features. For these features, we further propose a multi-layer dynamic-static aggregation strategy, which models the views-occlusion and globally completes information based on the multi-view relationships.
    \item We demonstrate that InsertNeRF achieves state-of-the-art performance with extensive generalization experiments by integrating the modules into the vanilla NeRF. Furthermore, we show the significant potential of our modules in various NeRF-like {systems}, {such as mip-NeRF~\citep{barron2021mip}, NeRF++~\citep{zhang2020nerf++}}, as shown in~\Cref{fig1}a, and in task with sparse inputs.
\end{itemize}

\section{Related Works}
\subsection{Generalizable Neural Radiance Fields}
Neural Radiance Fields (NeRF) by~\citep{mildenhall2021nerf} and its subsequent derivatives~\citep{barron2022mip, isaac2023exact, bao2023and} have gained momentum and are capable of producing realistic images. However, a significant drawback is the need to {retrain} them for every new scene, which is not efficient in real-world applications. Recent works by~\citep{wang2021ibrnet, wang2022attention} introduce Generalizable Neural Radiance Fields that can represent multiple scenes, regardless of whether they are in the training set. To achieve this, many studies have focused on understanding the relationships between reference views and refining NeRF's sampling-rendering mechanism. For instance, NeuRay~\citep{liu2022neural} and GeoNeRF~\citep{johari2022geonerf} use pre-generated depth maps or cost volumes as prior to alleviate occlusion issues. On the other hand, IBRNet~\citep{wang2021ibrnet} and GNT~\citep{wang2022attention} implicitly capture these relationships through MLPs or transformers. Regarding the sampling-rendering process, most works~\citep{xu2023wavenerf, suhail2022generalizable,wang2021ibrnet,zhu2023caesarnerf} utilize the transformer-based architectures to aggregate the sampling point features and replace traditional volume rendering with a learnable technique. However, a common limitation is that most of these methods replace NeRF's network with transformers, making it challenging to apply to NeRF derivatives and leading to increased computational complexity. Our research aims to address this by instilling generalizability into NeRF-like systems with scene-related weights while preserving its original framework and efficiency. 

\subsection{Hypernetworks}
The hypernetwork~\citep{ha2016hypernetworks, chauhan2023brief}, often abbreviated as hypernet, is invented to generate weights for a target neural network. Unlike traditional networks that require training from scratch, hypernets offer enhanced generalization and flexibility by adaptively parameterizing the target network~\citep{alaluf2022hyperstyle, yang2022dynamic, li2020dhp}. Leveraging these benefits, hypernets have found applications in various domains including few-shot learning~\citep{li2020dhp}, continual learning~\citep{von2019continual}, computer vision~\citep{alaluf2022hyperstyle}, etc. In the realm of NeRF, there have been efforts to incorporate hypernets to inform the training of the rendering process. For instance, \citep{chiang2022stylizing} propose to train a hypernet using style image features for style transfer, while~\citep{zimny2022points2nerf} employ encoded point-cloud features for volume rendering. On a related note, \citep{peng2023representing} utilize a dynamic MLP mapping technique to create volumetric videos and \citep{kania2023hypernerfgan} use a hypernet for 3D-aware NeRF GAN. In our work, instead of using the hypernet in NeRF framework directly, we introduce a plug-and-play HyperNet module, with a focus on providing reference scene knowledge to enable generalization to new scenarios.

\section{Method}
\subsection{Background}

\textbf{Neural Radiance Fields.} Neural radiance fields (NeRF)~\citep{mildenhall2021nerf} is a neural representation of scenes. It employs MLPs to map {a} 3D location $\bm{x} \in \mathbb{R} ^{3} $ and viewing direction $\bm{d} \in \mathbb{S} ^{2} $ to {an} emitted color $\bm c \in \left [ 0,1 \right ]^{3}$ and a volume density $\sigma \in \left [ 0,\infty \right ) $, which can be formalized as:
\begin{equation}
    \mathcal{F}(\bm x, \bm d ; \bm \Theta) \ \ \mapsto \ \  \left ( \bm c, \sigma \right ), \label{eq1}
\end{equation}
where $\mathcal{F}$ is the MLPs, and $\bm \Theta$ is the set of learnable parameters of NeRF.
Note that $\mathcal{F}$ can be further split into an appearance part $\mathcal{F}_{app}$ and a geometry part $\mathcal{F}_{geo}$ for the view-dependent attribute $\bm c$ and view-invariant attribute $\sigma$, respectively~\citep{zhang2023transforming}.

\textbf{Volume Rendering.} Given a ray in a NeRF, $\bm r \left ( t \right)= \bm{o}+{t}\bm{d}$, where $\bm{o}$ is the camera center and $\bm{d}$ is the ray's unit direction vector, we sample $K$ points, $\left \{ \bm r\left ( {t}_{i}\right ) |i=1,...,K \right \} $, along the ray and predict their color values $\bm c_{i}$ and volume densities $\sigma_{i}$. The ray's color is then calculated by:
\begin{equation}
\begin{gathered}
    \hat{C} \left ( \bm r \right ) = \sum_{i=1}^{K} w_{i} \bm c_{i}, \ \ \text{ where }\ \
     w_{i} = \exp\left ( -\sum_{j=1}^{i-1}\sigma_{j}\delta_{j}\right ) \left ( 1-\exp\left ( -\sigma_{i}\delta_{i} \right )  \right ),
\end{gathered}
\label{eq2}
\end{equation}
where $\delta_i$ is the distance between adjacent samples, and $w_{i}$ is considered to be the hitting probability or the weight of the $i$-th sampling point~\citep{liu2022neural}.

\textbf{Generalizable NeRF.} Given $N$ reference scene views with known camera poses $\left \{ \bm I_{n}, \bm P_n\right \}_{n=1}^N $, the goal of GNeRF is to synthesize a target {novel view $\bm I_T$} based on these reference views, even for scenes not observed in the training set, thereby achieving generalizability. Current works~\citep{wang2021ibrnet, liu2022neural, wang2022attention} primarily focus on aggregating features along with the ray $\bm r \left ( t \right)$ from multiple reference views. The overall process can be outlined as:
\begin{equation}
\mathcal{F}_{\text{sample}}\left(\left\{\mathcal{F}_{\text{view}} \left( \left\{\bm F_n\left( \Pi_n(\bm r(t_i))\right)\right\}_{n=1}^N\right)\right\}_{i=1}^K\right) \ \ \mapsto \ \ \bm (\bm c,\sigma).
\label{eq3}
\end{equation}
Here, $\Pi_n(\bm x)$ projects $\bm x$ onto $\bm I_n$, and $\bm F_n(\bm z)$ queries the corresponding feature vectors according to the projected points in {$n$-th reference}. $\mathcal{F}_{\text{view}}$ and $\mathcal{F}_{\text{sample}}$ specifically denote the aggregation of multi-view features and the accumulation of multiple sampling point features along the ray. These aggregations are often carried out using common techniques such as MLPs and transformers.

\subsection{InsertNeRF}\label{InsertNeRF}

\begin{figure}[t]
\begin{center}
\includegraphics[width=1\textwidth]{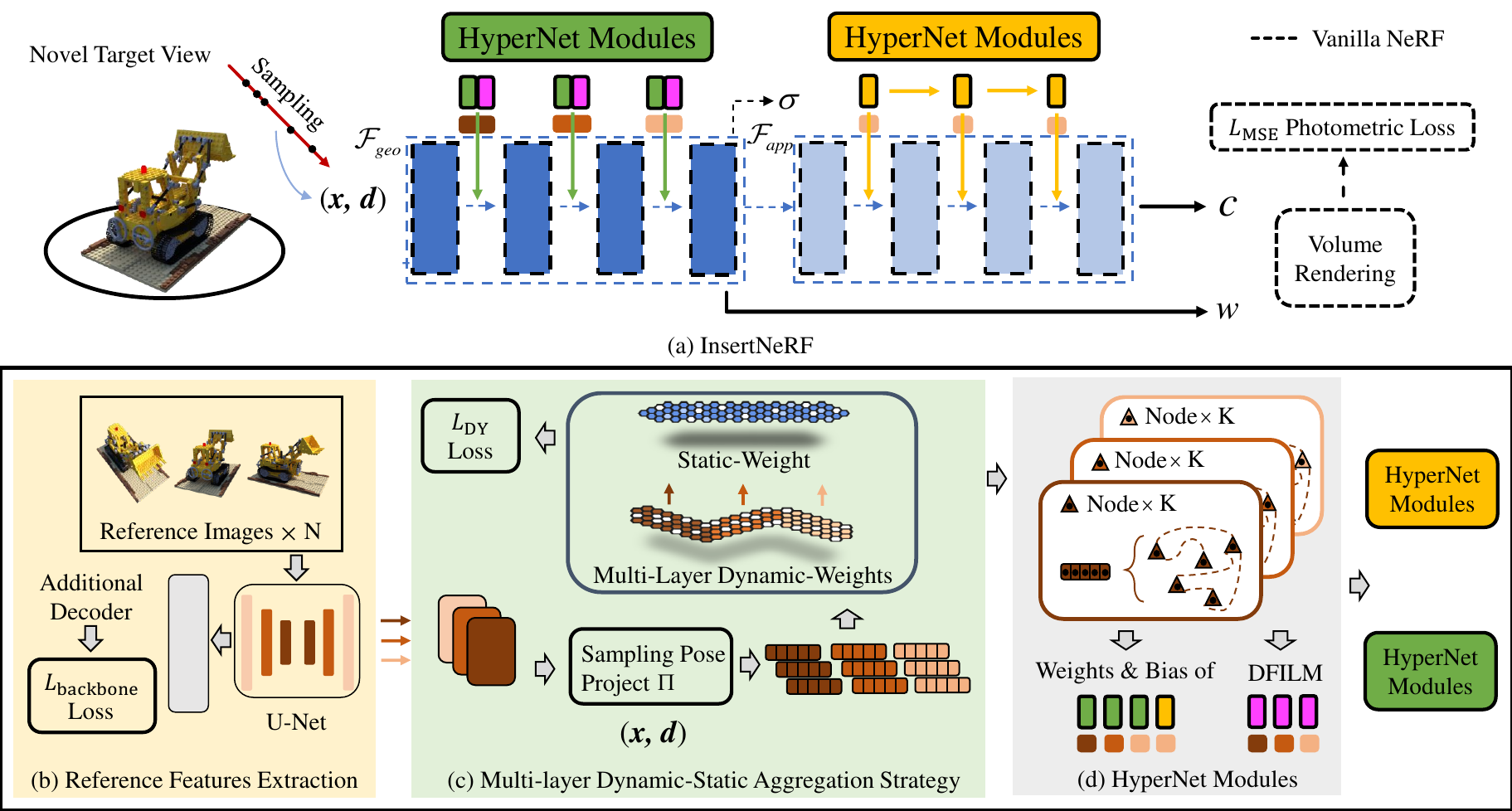}  
\end{center}
\vspace{-6pt}
\caption{
Overview of InsertNeRF. (a) Within the NeRF framework, two types of HyperNet modules are inserted into $\mathcal{F}_{geo}$ and $\mathcal{F}_{app}$. The HyperNet modules begin by (b) extracting features among multiple ($N$) reference images, and (c) using a multi-layer dynamic-static aggregation strategy to aggregate the scene representations. Based on these scene representations and specially designed sampling-aware filters, (d) we develop dynamic MLPs and activation functions to guide the weights and instill generalizability {into vanilla NeRF}. Finally, (a)standard volume rendering is performed.}

\vspace{-7pt}
\label{fig2}
\end{figure}

We introduce InsertNeRF, a novel paradigm that instills generalizability into the NeRF framework, as illustrated in~\Cref{fig2}. While this method can be adapted to a variety of NeRF-based systems~(\Cref{sec:more-frameworks}), we focus on its application on the vanilla NeRF in this section. 


\textbf{Overview.} 
InsertNeRF achieves generalizability by inserting multiple HyperNet modules into NeRF. These modules dynamically generate scene-specific weights for NeRF that are tailored to specific {reference} scene, denoted by $\bm \Omega_T$. Based on it, by incorporating $\bm \Omega_T$ into~\Cref{eq1} and combining $\bm \Theta$ as well as $\bm \Omega_T$, NeRF's implicit representation (or weights) gains the generalizability across multi-scenes, which is explained in~\Cref{D3}. Specifically, it can be described as follows:
\begin{align}
    \begin{split}\label{eq4}
        &\mathcal{F}(\bm x, \bm d ; \bm \Theta, \bm\Omega_T) \ \ \mapsto \ \  \left ( \bm c, w \right ), \\
    \text{ where }\ \bm\Omega_T = \ & \text{HyperNet} \left ( \left\{\mathcal{F}_{\text{view}}\left(\left\{\bm F_n\left( \Pi_n(\bm r(t_i)) \right)\right\}_{n=1}^N\right)\right\}_{i=1}^K\right ).
    \end{split}
\end{align}
Comparing~\Cref{eq4} to~\Cref{eq3}, the key to InsertNeRF is the newly introduced architectures with dynamic weights $\bm\Omega_T$, guided by the HyperNet modules based on specific reference inputs. The process begins with reference features extraction~(\Cref{Extraction}), then a multi-layer dynamic-static aggregation strategy is employed to fuse reference features from multi-views into scene features~(\Cref{Multi-layer}). Subsequently, these aggregated scene features are used to adaptively generate NeRF's sampling-aware weights via the HyperNet modules, which consist of sampling-aware filters, dynamic MLPs and dynamic activation functions~(\Cref{HyperNet}). These novel HyperNet modules are inserted before each MLP layer in the original NeRF, serving as an enhancement to the original MLP layers.

A notable aspect of InsertNeRF is its ability to directly calculate the hitting probability $w_i$ in~\Cref{eq2} for volume rendering, rather than simply outputting the volume density from $\mathcal{F}_{geo}$. This capability stems from the implicit modeling of the relationships between spatial points and the advantage of using multi-scale features. By combining $\mathcal{F}_{geo}$ with $\mathcal{F}_{app}$,
the entire pipeline is trained end-to-end. Our unique design not only leads to superior rendering performance in GNeRF but also offers improved computational efficiency compared to transformer-based structures~\citep{wang2022attention}.



\subsubsection{Reference Features Extraction}\label{Extraction}

In the exploration of reference images, generalizable methods often combine U-Net~\citep{ronneberger2015u} and ResNet~\citep{he2016deep} to extract local dense feature maps. These have proven effective in dealing with occlusion problems~\citep{liu2022neural}. Yet, there is a risk that an overemphasis on local dense features might neglect global features, which are key to occlusion completion {and global inference}~\citep{iizuka2017globally}.
In our work, we take advantage of the spatial representation capabilities of multi-scale features to model complex geometry and detailed appearance. Specifically, we bring in global-local features to successively  update {the Hypernet module's weights for $\mathcal F_{geo}$ and dense feature for $\mathcal F_{app}$}. Here, geometry requires multi-scale information to deduce occluded portions, while appearance concentrates on dense fine-grained details.
This process begins with multi-scale features $\bm F_{l,n}$ from U-Net for each reference input $\bm I_n$, and can be expressed as:
\begin{equation}
\bm F_{l,n}\in \mathbb {R}^{\frac{W}{2^{l+1}}\times \frac{H}{2^{l+1}} \times C_{l}}, \quad  l=  2,1,0; \ n=1,\cdots, N.   \label{eq5}
\end{equation}
Here, $W\times H$ defines the image resolution, and $C_l$ is the number of channels. During feature upsampling (as $l$ decreases), we output each layer's features, transitioning from global to local.

\subsubsection{Multi-Layer Dynamic-Static Aggregation Strategy}\label{Multi-layer}

Following the feature extraction, the next essential step is the aggregation of scene features. This is not only foundational for scene generalizability but also significantly impacts the effectiveness of the HyperNet modules. 
Most existing techniques focus primarily on preserving local geometry and appearance consistency, often employing visibility to model occlusions. A straightforward approach is to deduce the view-weight based on differences between reference and target views~\citep{wang2021ibrnet}. {We refer to it} as \textit{static weight}, denoted by {$M^{\text{ST}}\in\mathbb {R}^{B\times K \times N}$}, where $B$ represents the batch size, and it assigns higher weights to closer views in a fixed manner. However, it may be unreliable as it overlooks the correlation among the features. To remedy this, we introduce a dynamic prediction of multi-layer weights based on multi-scale features, involving a blend of Maxpool-MLPs and Softmax layers, termed \textit{dynamic weights} and denoted by $M_l^{\text{DY}}\in\mathbb {R}^{B\times K \times N}$. 
Our approach hence adopts a \textit{dynamic-static} aggregation strategy for more nuanced multi-view scene feature aggregation.

Formally, given the corresponding features $\bm F_l\in\mathbb{R}^{B\times K\times N\times d_l}$ of $B\times K$ points in the space, where $d_l$ is the latent feature dimension, we calculate the weighted means and variances as  $\bm\mu_l=\mathbb{E}_n\left[\bm F_{l}\odot M_l^{\text{DY}}\right]\in\mathbb{R}^{B\times K\times d_l}$ and $\bm v_l=\mathbb{V}_n\left[\bm F_{l}\odot M_l^{\text{DY}}\right]\in\mathbb{R}^{B\times K\times d_l}$, respectively. After concatenating $\bm F_l$ for each reference view with $\bm \mu_l$ and $\bm v_l$ and halvely projecting its dimension, denoted as $\widetilde{\bm F}_l\in\mathbb{R}^{B\times K\times N\times d_l/2}$, it is applied to the static weight to obtain $\widetilde{\bm\mu}_l=\mathbb{E}_n\left[\widetilde{\bm F}_{l}\odot M^{\text{ST}}\right]\in\mathbb{R}^{B\times K\times d_l/2}$ and $\widetilde{\bm v}_l=\mathbb{V}_n\left[\widetilde{\bm F}_{l}\odot M^{\text{ST}}\right]\in\mathbb{R}^{B\times K\times d_l/2}$. With $\bm F_l^{\text{max}}\in\mathbb{R}^{B\times K\times d_l}$ representing the maximum features among all the reference views, and by concatenating $\widetilde{\bm \mu}_l$ and $\widetilde{\bm v}_l$, and adding it to $\bm F_l^{\text{max}}$, we accomplish the feature aggregation phase $\mathcal{F}_{\text{view}}$ in~\Cref{eq4}.\footnote{The advantages of this strategy in comparison with~\cite{wang2021ibrnet} are discussed in the~\Cref{D1}.}

The use of global-local dynamic weights leads to a significant enhancement in edge sharpness and the thorough completion of detail in the depth rendering images, as evidenced in~\Cref{fig1}b. Note that unlike static weights, dynamic weights are guided by the relationships between multi-scale reference features and are learned with auxiliary supervision (\Cref{Loss}).


\subsubsection{HyperNet Modules}\label{HyperNet}

We now turn our attention to the HyperNet modules, the core element of InsertNeRF, integrated within both $\mathcal F_{geo}$ and $\mathcal F_{app}$. These modules are composed of three basic components: sampling-aware filters, dynamic MLPs (D-MLP), and dynamic activation functions.

\textbf{Sampling-aware Filter.} 
Unlike traditional hypernetworks, where reference features are generally stable, those based on {pose-related} epipolar geometric constraints in GNeRF are noisy. This noise complicates their direct use for weights generation. To address this challenge, we introduce a sampling-aware filter that seeks to implicitly find correlations between inter-samples and reduce noise within the reference features through graph reasoning. Specifically, following the aggregation phase $\mathcal F_\text{view}$, each aggregated point-feature is regarded as a node within a graph structure. The relationships between these $K$ points are then modeled using graph convolutions, formulated as:
\begin{equation}
\bm {H}_{l} = 
\left ( \bm {I} - \bm {A}_{l}\right ) \bm F_{\text{view}} \bm {W}^{a}_{l},
\label{eq6}
\end{equation}
where $ \bm F_{\text{view}} \in \mathbb {R}^{B\times K \times d_{l}}$ denotes the aggregated {$K$ point-features} after $\mathcal{F}_{\text{view}}$, and $\bm {A}_{l}$ and $\bm {W}^a_{l}$ represent the $K\times K$ node adjacency matrix and the learnable {state update function}, respectively. $\bm {I}$ here denotes the identity matrix. This specific graph structure helps filter out noise by {state-updating}, enabling the network to concentrate on key features more effectively. Additionally, for intricate tiny structures, we adopt an approach inspired by~\cite{chen2019graph}, where linear layers across different dimensions are utilized instead of standard matrix multiplications within the graph convolutions.

\textbf{Dynamic MLP.} 
Using the filtered features $\bm {H}_{l}$, the HyperNet module is designed to generate corresponding Weight$_{\bm {H}_{l}}$ and Bias$_{\bm {H}_{l}}$ within specific MLPs. This instills scene-awareness into vanilla NeRF, ensuring compatibility with $\bm {F}_\text{input}$, the output of the previous layer in the original NeRF framework. To enhance efficiency, these {MLPs} are integrated within the sampling-aware filter.

\textbf{Dynamic Activation Function.} 
Activation functions plays an essential role in the NeRF framework~\citep{sitzmann2020implicit}. Traditional options, such as the ReLU function, may struggle with detail rendering and hinder the performance of D-MLPs due to their static nature. To address this, we introduce a dynamic activation function. This function adaptively activates features in accordance with the unique characteristics of a given scene. Inspired by~\cite{perez2018film}, we propose the Dynamic Feature-wise Linear Modulation (DFiLM), in which the frequencies (Freq$_{\bm {H}_{l}}$) and phase-shifts (Shift$_{\bm {H}_{l}}$) are dynamically determined from $\bm {H}_{l}$, allowing for more responsive activation.

The entire MLP-Block, including both the D-MLP and the activation function, can be expressed as:
\begin{equation}
\bm {F}_\text{output} =\text{Shift}_{\bm {H}_{l}} (\text{Freq}_{\bm {H}_{l}}(\text{Weight}_{\bm {H}_{l}}\times \bm {F}_\text{input} +\text{Bias}_{\bm {H}_{l}}
)) ,\label{eq7}
\end{equation}
To insert the HyperNet modules into the NeRF framework, $\bm F_{\text{output}}$ is subsequently fed into an original NeRF's MLP layer for the final result. This yields superior performance, as validated through experimental results. We remark that the parameters are not shared among the HyperNet modules. Moreover, their compact structures ensure that the impact on rendering efficiency is negligible.

\textbf{HyperNet Modules in $\mathcal{F}_{geo}$ and $\mathcal{F}_{app}$.} In vanilla NeRF, $\mathcal{F}_{geo}$ and $\mathcal{F}_{app}$ serve distinct purposes but employ similar MLP structures, albeit with varying complexities. $\mathcal{F}_{geo}$ focuses on geometric properties, whereas $\mathcal{F}_{app}$ encodes view-dependent features using a smooth BRDF prior for surface reflectance. This smoothness can be facilitated by progressively exploiting guided scene features, {along with a reduction in both MLP parameters and activation functions for variable $\bm d$}~\citep{zhang2020nerf++}. Recognizing this need, we propose a modified HyperNet module architecture specifically for $\mathcal{F}_{app}$. Our design employs a progressive guidance mechanism within $\mathcal{F}_{app}$, incorporating multiple parallel dynamic branches into the NeRF framework. The weights of the D-MLP in each branch are progressively generated from the preceding branch, enabling the capture of reference features at different levels for complex appearance modeling. Finally, the results of all branches are summed and used as input to the original MLP for predicting the RGB value. In accordance with our analysis, the DFiLM is not used in $\mathcal{F}_{app}$, setting it apart from other elements in the architecture. [\Cref{C1}]

\subsection{Loss Functions}\label{Loss}

The InsertNeRF pipeline is trained end-to-end utilizing three carefully designed loss functions.

\textbf{Photometric loss.} First, we employ the photometric loss in NeRF~\citep{mildenhall2021nerf}, i.e., the Mean Square Error (MSE) between the rendered and true pixel colors:
\begin{equation}
\mathcal L_{\text{MSE}} = \sum_{\bm{r}\in {\mathcal{R}}} \left \| \hat{C}(\bm{r}) - C(\bm{r})  \right \|_{2}^{2}, \label{eq11}
\end{equation}
where $\mathcal{R}$ is the set of rays in a batch, and $C(\bm{r})$ is the ground-truth RGB color for ray $\bm{r}\in \mathcal{R}$. 

\textbf{Backbone loss.} During end-to-end training, optimizing the feature extraction without additional guidance poses considerable challenges. To address this, we draw inspiration from auto-encoding~\citep{kingma2013auto}. By adding an {additional} upsampling layer and a small decoder (used exclusively for loss computation), we seek to reconstruct reference images from encoded features. The original images serve as supervision, and we refer to this particular loss term as $\mathcal{L}_{\text{backbone}}$.

\textbf{Dynamic weights loss.} Initiating the learning of dynamic weights from scratch introduces difficulties in understanding the connections among multi-scale features. To tackle this issue, we introduce an auxiliary supervision to encompass global-local information. Specifically,  we let $C^{\text{ref}}_n (\bm r)\in \mathbb {R}^{B\times K \times N \times 3}$ represent the ground-truth RGB values in corresponding reference images for $K$ points in ray $\bm r$ within a batch $\mathcal{R}$. We compute $\bm c'_{i}  = \sum_{n,l, \bm r\in\mathcal{R}} C^{\text{ref}}_n (\bm r)\odot M_l^{\text{DY}}$, the weighted sum of these RGB values by dynamic weights. Utilizing $\bm c'_i$,  ${\hat{C}}'\left ( \bm{r} \right )$ is subsequently calculated according to~\Cref{eq2}, and supervised by the true color $C(\bm r)$. We designate this loss term as $\mathcal L_\text {DY}$.

We formulate our final loss function as
\begin{equation}
    \mathcal{L} = \mathcal L_{\text{MSE}} + \lambda_1 \mathcal{L}_{\text{backbone}} + \lambda_2\mathcal L_\text {DY}
\end{equation}
where $\lambda_1$ and $\lambda_2$ are hyperparameters controlling the relative importance of these terms.

\begin{table*}
    \centering
        \caption{{Comparisons of InsertNeRF against SOTA methods with Setting I.} 
        }
        \vspace{-.1in}
    \scalebox{0.80}{
    \setlength{\tabcolsep}{1mm}{
    \begin{tabular}{l|ccc|ccc|ccc}
        \toprule[2pt]
            \multirow{2}{*}{Methods}& \multicolumn{3}{c|}{NeRF Synthetic} & \multicolumn{3}{c|}{LLFF} & \multicolumn{3}{c}{DTU}\\
          & PSNR$\uparrow$ & SSIM$\uparrow$ & LPIPS$\downarrow$ & PSNR$\uparrow$ & SSIM$\uparrow$ & LPIPS$\downarrow$& PSNR$\uparrow$ & SSIM$\uparrow$ & LPIPS$\downarrow$\\
        \midrule
        PixelNeRF (CVPR2021) & 22.65& 0.808& 0.202& 18.66& 0.588&0.463 &19.40 &0.463 &0.447\\
        MVSNeRF (ICCV2021)&25.15& 0.853& 0.159& 21.18& 0.691&0.301 &23.83 &0.723 &0.286\\
        IBRNet (CVPR2021)&26.73& 0.908& 0.101& 25.17& 0.813&0.200 &25.76 &0.861 &0.173\\
        ContraNeRF (CVPR2023)& - & - & - & 25.44& \underline{0.842}&\underline{0.178} &27.69 &0.904 &\underline{0.129}\\
        GeoNeRF{$^\dag$} (CVPR2022) & 28.33& \underline{0.938}& \underline{0.087}& 25.44& 0.839 &0.180 & -  & -  & -\\
        WaveNeRF{$^\dag$} (ICCV2023)&26.12& 0.918& 0.113& 24.28& 0.794 &0.212 & -  & - & -\\
        NeuRay(CVPR2022)& \underline{28.92}& 0.920& 0.096& \underline{25.85}& 0.832 &0.190 &\underline{28.30} &\underline{0.907} &0.130\\
        \midrule
        InsertNeRF (\textbf{Ours})& \textbf{30.35}& \textbf{0.938}& \textbf{0.065}& \textbf{26.44}& \textbf{0.844}&\textbf{0.169} &\textbf{29.75} &\textbf{0.925} &\textbf{0.077} \\
        \bottomrule[2pt]
    \end{tabular}}}
    \label{tab1}
    \vspace{-.1in}
\end{table*}

\begin{table}[t]
  \begin{minipage}{0.55\textwidth}
    \centering
    \caption{{Comparisons and ablations with Setting II.}}
    \label{tab2}
     \vspace{-.1in}
    \scalebox{0.63}{
    \begin{tabular}{l|ccc|ccc}
        \toprule[2pt]
            \multirow{2}{*}{Methods}& \multicolumn{3}{c|}{NeRF Synthetic} & \multicolumn{3}{c}{LLFF} \\
          & PSNR$\uparrow$ & SSIM$\uparrow$ & LPIPS$\downarrow$ & PSNR$\uparrow$ & SSIM$\uparrow$ & LPIPS$\downarrow$ \\
        \midrule
        GNT (ICLR2023)& 27.29& \textbf{0.937}& 0.056& 25.59& 0.858&0.128\\
        \midrule
        Baseline (NeRF)& 7.29 &0.512 &0.690 &11.46 &0.328 &0.582\\
        NeRF with HyperNetwork  & 25.86 &0.902 &0.081 &24.25 &0.793 &0.177\\
        InsertNeRF w/o MLDS &25.12 &0.896 &0.098 &24.41 &0.814 &0.156\\
        InsertNeRF (\textbf{Ours})& \textbf{27.57}& \underline{0.936}& \textbf{0.056}& \textbf{25.68}& \textbf{0.861}&\textbf{0.126} \\
        \bottomrule[2pt]
    \end{tabular}}
  \end{minipage}%
  \hspace{.2in}
  \begin{minipage}{0.4\textwidth}
    \centering
    \caption{{Results with sparse inputs.}}
    \label{tab3}
        \vspace{-.1in}
    \scalebox{0.62}{
    \begin{tabular}{l|ccc}
        \toprule[2pt]
            \multirow{2}{*}{Methods}&  \multicolumn{3}{c}{3-view} \\
          & PSNR$\uparrow$ & SSIM$\uparrow$ & LPIPS$\downarrow$  \\
        \midrule
        DietNeRF (ICCV 2021) & 14.94 & 0.370& 0.496\\
        RegNeRF (CVPR 2022) & 19.08& 0.587& 0.336\\
        GeCoNeRF (ICML 2023) & 18.77& 0.596 & 0.338\\
        FreeNeRF (CVPR 2023) & \textbf{19.63}& 0.612& \textbf{0.308}\\
        InsertNeRF (w/o retrain)& \underline{19.41}& \textbf{0.618}&\underline{0.330} \\
        \bottomrule[2pt]
    \end{tabular}}
  \end{minipage}
  \vspace{-.2in}
\end{table}

\section{Experiments}


We conduct comparative experiments with state-of-the-art (SOTA) methods across different settings on mainstream datasets. Additionally, we validate the effectiveness of the proposed paradigm in the context of {derivative} NeRF-like {systems} generalizations and tasks involving sparse inputs.


\subsection{Experimental Protocol and Settings}\label{settings}

Following IBRNet~\citep{wang2021ibrnet}, GNeRF exploits the target-reference pairs sampling strategy during both the training and inference phases. Here, reference views are selected from a set of nearby views surrounding the target view. Specifically, $N$ reference views are chosen from a pool of $P \times N$ ($P\ge 1$) neighboring views of target, ensuring that the target view is excluded from the reference views. During the evaluation phase, we conduct evaluations using three metrics: PSNR, SSIM, and LPIPS, on well-established datasets such as NeRF Synthetic, LLFF, and DTU. More training and inference details are provided in the~\Cref{A} and Algorithm in the~\Cref{C2}.


In our experiments, we follow two GNeRF settings of existing methods:

\textbf{Setting I.} Following NeuRay~\citep{liu2022neural}, we use three types of training datasets for training GNeRF, including three forward-facing datasets, the synthetic Google Scanned Object dataset and the DTU dataset. {Note that we only select training scenes in the DTU dataset, excluding the four evaluation scenes.} Following their setting in the experiments, we set $N=8$.

\textbf{Setting II.} Following GNT~\citep{wang2022attention}, we train GNeRF using three forward-facing datasets and the Google Scanned Object dataset. Unlike Setting I, the DTU dataset is not used for either training or evaluation. In addition, we set $N=10$ in this setting.



\subsection{Comparative Experiments}

We evaluate InsertNeRF for its generalization based on the vanilla NeRF framework, comparing its performance with SOTA methods under two GNeRF settings. Through extensive {quantitative and qualitative} experiments, we explore the advantages of our approach, even with {fewer references}.

\begin{figure}[t]
\begin{center}{
\includegraphics[width=1\textwidth]{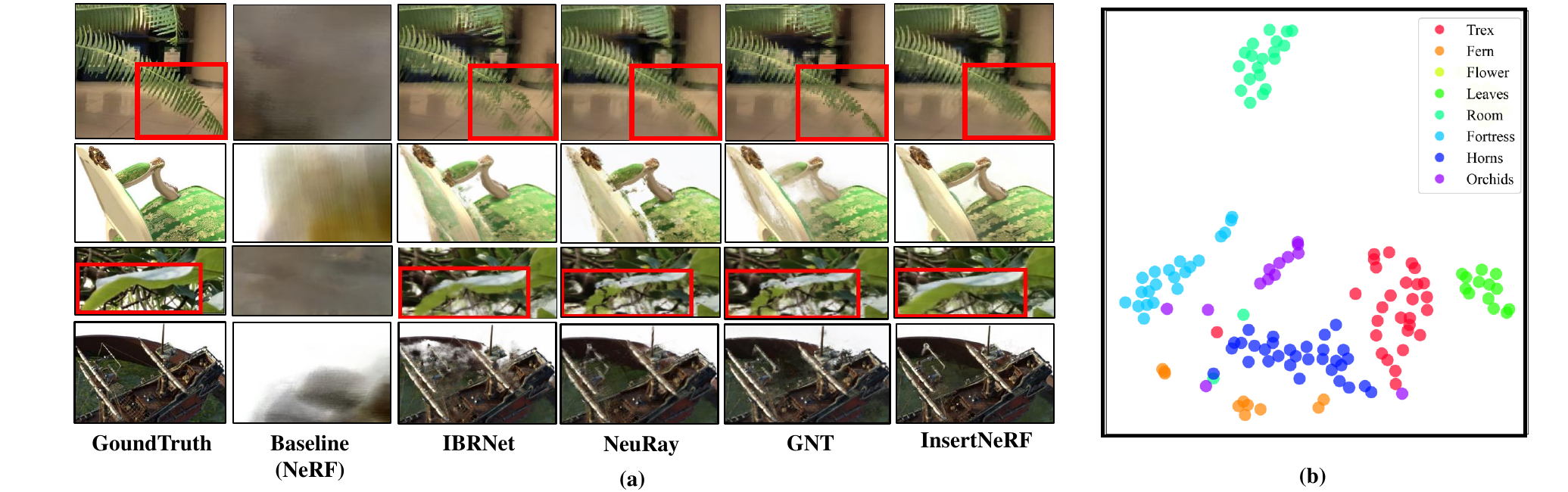} 
\vspace{-3mm}
\caption{(a) Qualitative comparisons of InsertNeRF against SOTA methods. (b) {A t-SNE plot of the scene-specific representations from our HyperNet modules. More analysis in the~\Cref{D2}} 
\label{fig5}}
\vspace{-17pt}}
\end{center}
\end{figure}

\begin{table}[t]
  \begin{minipage}{0.40\textwidth}
    \centering
    \caption{HyperNet modules ablations.}
    \label{tab4}
     \vspace{-.1in}
    \scalebox{0.70}{
    \begin{tabular}{l|ccc}
        \toprule[2pt]
            \multirow{2}{*}{Methods}&  \multicolumn{3}{c}{LLFF} \\
          & PSNR$\uparrow$ & SSIM$\uparrow$ & LPIPS$\downarrow$  \\
        \midrule
        w/o D-MLP &23.33  &0.774 & 0.198 \\
        w/o Sampling Filter & 24.67 & 0.815& 0.158\\
        w/o DFiLM & 25.04& 0.832& 0.152\\
        \midrule
        w/o original MLP & 25.44& 0.848& 0.131\\
        InsertNeRF (\textbf{Ours})& \textbf{25.68}& \textbf{0.861}&\textbf{0.126} \\
        \bottomrule[2pt]
    \end{tabular}}
  \end{minipage}%
  \hspace{.1in}
  \begin{minipage}{0.55\textwidth}
    \centering
     \caption{MLDS aggregation strategy ablations.}
    \label{tab5}
        \vspace{-.1in}
    \scalebox{0.70}{
    \setlength{\tabcolsep}{1mm}{
    \begin{tabular}{ccccc|ccc}
        \toprule[2pt]
             Static- & Dynamic-& Auxiliary-& Multi-  & Single- & \multicolumn{3}{c}{LLFF} \\
            Weight & Weight& Supervision & Layers  & Layer & PSNR$\uparrow$ & SSIM$\uparrow$ & LPIPS$\downarrow$  \\
        \midrule
        \checkmark & &  &\checkmark & &  24.88& 0.827& 0.154\\
        \checkmark&\checkmark & &\checkmark & &25.55 &0.851 &0.128 \\
        &\checkmark &\checkmark &\checkmark & & 25.53& 0.850& 0.131\\
        \midrule
        \checkmark & \checkmark  &\checkmark &  &\checkmark & 25.15& 0.838& 0.139\\
        \checkmark& \checkmark& \checkmark&\checkmark & &\textbf{25.68}& 
        \textbf{0.861}&\textbf{0.126} \\
        \bottomrule[2pt]
    \end{tabular}}}
  \end{minipage}
  \vspace{-.25in}
\end{table}

\renewcommand{\thefootnote}{\dag} 
\footnotetext{GeoNeRF~\citep{johari2022geonerf} and WaveNeRF~\citep{xu2023wavenerf} are trained on original rectified 
images and evaluated on the distinct scenes with us in the DTU dataset.}
\renewcommand{\thefootnote}{\arabic{footnote}} 

\textbf{Quantitative comparisons.} We present quantitative comparisons with SOTA methods under Setting~I and Setting~II, as reported in~\Cref{tab1} and~\Cref{tab2}. For Setting I, the quantitative comparisons in~\Cref{tab1} display our model’s competitive results in evaluation datasets, with significant improvements in PSNR, SSIM and LPIPS in comparison to existing SOTA methods. Specifically,  PSNR and LPIPS exhibit substantial enhancements by $\sim$1.16dB$\ \uparrow$ and $\sim$23.6$\%\downarrow$ respectively. For Setting~II, InsertNeRF consistently outperforms the SOTA method~\citep{wang2022attention}, as substantiated by the results in~\Cref{tab2}. We observe that these improvements become even more pronounced with fewer reference images, alongside higher efficiency, as demonstrated in subsequent sections.


\textbf{Qualitative comparisons.} \Cref{fig1} and~\Cref{fig5} {(a)} show the qualitative performances of our method against baseline and SOTA methods. InsertNeRF achieves improved geometric fidelity and clear edges, attributable to the completion capability of global features and the modeling of sample spatial relationships from graph structures. For more analysis and results, please refer to the~\Cref{E4}.

\subsection{Ablation Studies} 

\begin{wrapfigure}{r}{6.5cm}
\vspace{-6mm}
\includegraphics[width=0.45\textwidth]{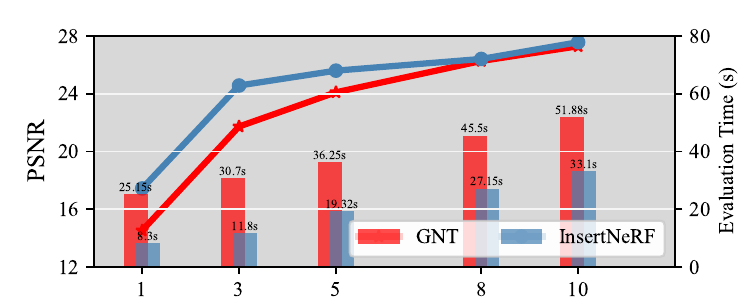} 
\vspace{-2mm}
\caption{Performance and efficiency under different input-number $N$ on NeRF Synthetic.}
\label{fig6}
\vspace{-2mm}
\end{wrapfigure}
In~\Cref{tab2}, we analyze the core components of our method. The findings underscore the vital role of HyperNet modules in enhancing rendering performance, as further evidenced in~\Cref{fig5} (b) that they instill scene-specific capabilities into NeRF’s representation. Additionally, the multi-layer dynamic-static aggregation strategy proves to be essential. By integrating both modules, {our novel paradigm instills generalizability into the NeRF framework}, leading to a performance boost of approximately two to three times compared to the baseline model, {i.e., vanilla NeRF}. Additionally, we explore the underlying mechanisms driving the effectiveness of these components. More experiments, including single-scene setting, fine-tuning and ablation studies about $\mathcal{F}_{app}$ in the~\Cref{E}

\textbf{HyperNet modules.} \Cref{tab4} demonstrates that both the sampling-aware filters and dynamic activation functions are vital in the HyperNet modules, with the sampling-aware filters having a more substantial impact. This could be due to the need to consider relationships between sampled points in the rendering process, which implicitly models occlusions, as noted in~\cite{liu2022neural}. Solely using dynamic activation functions without D-MLP leads to a marked decline in performance, highlighting the essential role of MLPs in neural representation. Furthermore, using only the HyperNet modules and omitting the original NeRF's MLP layers results in inferior performance, reducing training stability.

\textbf{Multi-layer dynamic-static aggregation strategy.} In \Cref{tab5}, ablation studies reveal the significance of dynamic-static weights and multi-layer features. Using only {dynamic weights} appears more effective than static weight, likely because they are adaptively generated to suit different scene features. The auxiliary supervision for dynamic weights and multi-layer global-local features also play essential roles in aggregating multi-view features, underlining their importance in this strategy.

\textbf{Input number ($N$) and efficiency.} Since feature extraction is time-consuming, reducing the number of reference images substantially improves the training and inference efficiency of the network. \Cref{fig6}~illustrates the performance of InsertNeRF as the number of reference images ($N$) varies for training on NeRF Synthetic. In comparison to GNT~\citep{wang2022attention}, InsertNeRF consistently demonstrates superior rendering performance and inference efficiency. This success can be attributed to our novel generalization paradigm and the compact structures of the HyperNet modules.

\begin{table*}
    \centering
        \caption{Quantitative results of InsertNeRF and Insert-mip-NeRF on multi-scale NeRF Synthetic.}
        \vspace{-.1in}
    \scalebox{0.57}{
    \setlength{\tabcolsep}{2.80mm}{
    \begin{tabular}{l|cccc|cccc|cccc}
        \toprule[2pt]
            \multirow{2}{*}{Methods}& \multicolumn{4}{c|}{PSNR$\uparrow$ } & \multicolumn{4}{c|}{SSIM$\uparrow$ } & \multicolumn{4}{c}{LPIPS$\downarrow$}\\
          &Full Res.&1/2 Res.&1/4 Res.&1/8 Res. &Full Res.&1/2 Res.&1/4 Res.&1/8 Res. &Full Res.&1/2 Res.&1/4 Res.&1/8 Res. \\
        \midrule
        mip-NeRF&12.94& 13.03& 13.18& 13.33& 0.700&0.636 &0.563 &0.469 &0.424&0.460&0.470&0.530\\
        InsertNeRF&27.60& 28.58& 29.45& 29.85& 0.926&0.943 &0.960 &0.972 &0.066&0.054&0.045&0.036\\
        Insert-mip-NeRF&\textbf{28.15}& \textbf{29.17}& \textbf{30.22}& \textbf{30.62}& \textbf{0.935}&\textbf{0.951} &\textbf{0.966} &\textbf{0.977} &\textbf{0.056}&\textbf{0.045}&\textbf{0.037}&\textbf{0.029}\\
        \bottomrule[2pt]
    \end{tabular}}}
    \label{tab6}
    \vspace{-7.3mm}
\end{table*}

\subsection{Insert-NeRF-like Frameworks} \label{sec:more-frameworks}

Thanks to the plug-and-play advantage of the HyperNet modules, we extend the study of generalization to derived domains of NeRF, such as mip-NeRF~\citep{barron2021mip} and {NeRF++~\citep{zhang2020nerf++}}, areas that have rarely been discussed before. More details are provided in the~\Cref{A}.

\begin{wrapfigure}{}{7.0cm}
\vspace{-6mm}
\includegraphics[width=0.50\textwidth]{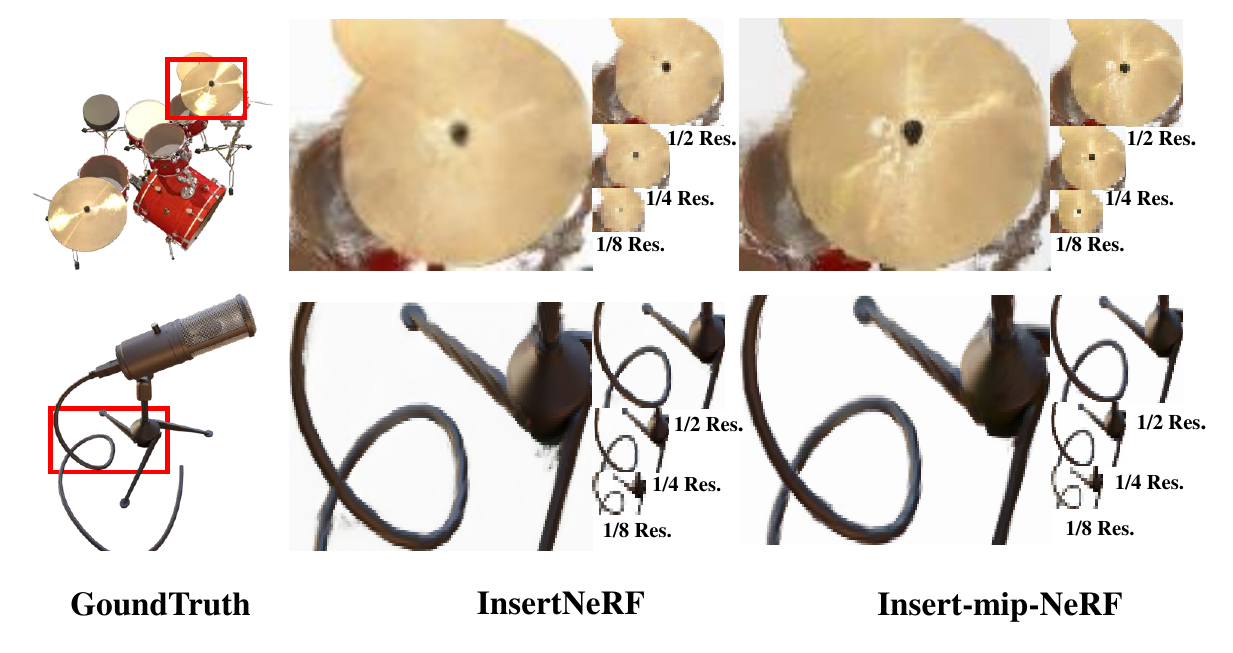} 
\caption{Qualitative results of Insert-mip-NeRF. Please refer to the~\Cref{E4} for more results.}
\vspace{-4mm}
\label{fig4}
\end{wrapfigure}

\textbf{Insert-mip-NeRF.} Mip-NeRF is a multi-scale NeRF-like model used to address the inherent aliasing of NeRF, a significant challenge for GNeRF. Unlike~\cite{huang2023local}, we explore how to instill generalizability into mip-NeRF, following its original setup. We report the qualitative and quantitative performance of mip-NeRF, InsertNeRF, and Insert-mip-NeRF on multi-scale NeRF Synthetic in a cross-scene generalization setting (see~\Cref{tab6},~\Cref{fig1} and~\Cref{fig4}). One can observe that incorporating the HyperNet modules not only enhances generalization for mip-NeRF but also addresses the inherent aliasing of InsertNeRF and improves the performance in the task of multi-scale rendering.

\textbf{Insert-NeRF++.} NeRF++, an unbounded NeRF. \Cref{fig1} and \Cref{fig13} depicts qualitative and quantitative rendering results of Insert-NeRF++. It is evident that our approach has successfully instilled generalizability into the NeRF++ framework, doubling its PSNR compared to the original. 

\textbf{Sparse Inputs.} Training NeRF with sparse inputs has become a notable focus recently~\citep{niemeyer2022regnerf, yang2023freenerf}. Unlike our nearby reference views setting (\Cref{settings}), this task often involves training from a limited number of fixed viewpoints to represent the entire scene. Under this setting, we relax constraints on selecting nearby viewpoints and uniformly select fixed sparse seen viewpoints to infer on arbitrary unseen viewpoints. Unlike existing works, our method trains on extensive auxiliary datasets, allowing us to represent the entire evaluation scene from sparse inputs without retraining (see~\Cref{tab3}). To ensure fairness, all scenes in evaluation are excluded in the training phase. In conclusion, InsertNeRF offers a novel insight that employs pre-training on auxiliary datasets to enhance representation capabilities with sparse inputs. We believe that, through fine-tuning on the evaluation scene and incorporating existing technologies like geometry and color regularization, our paradigm will achieve even better performance under sparse inputs.



\section{Conclusion}


We present InsertNeRF, a novel paradigm that {instills generalizability into NeRF systems.} Unlike popular transformer-based structures, our HyperNet modules are efficiently incorporated into the original NeRF-like framework, leveraging reference scene features to generate scene-specific network weights. {To achieve this}, we design a multi-layer dynamic-static feature aggregation strategy for extracting scene features from reference images and employ sampling-aware filters to explore relationships between sample points. Experiments on well-established datasets show that InsertNeRF and other Insert-NeRF-like frameworks can render high-quality images across different scenes without retraining. This offers insights for future works on: (i) generalization tasks for additional NeRF-like systems such as mip-NeRF 360; and (ii) sparse inputs tasks based on {auxiliary datasets}.

\section*{Acknowlefgements}
This work was supported in part by the National Natural Science Foundation of China under Grant 62276128, Grant 62192783, and Grant 62106100; in part by the Jiangsu Natural Science Foundation under Grant BK20221441; in part by the Young Elite Scientists Sponsorship Program by CAST under Grant 2023QNRC001; in part by the Collaborative Innovation Center of Novel Software Technology and Industrialization.

\bibliography{iclr2024_conference}

\begin{thebibliography}{64}
\providecommand{\natexlab}[1]{#1}
\providecommand{\url}[1]{\texttt{#1}}
\expandafter\ifx\csname urlstyle\endcsname\relax
  \providecommand{\doi}[1]{doi: #1}\else
  \providecommand{\doi}{doi: \begingroup \urlstyle{rm}\Url}\fi

\bibitem[Alaluf et~al.(2022)Alaluf, Tov, Mokady, Gal, and Bermano]{alaluf2022hyperstyle}
Yuval Alaluf, Omer Tov, Ron Mokady, Rinon Gal, and Amit Bermano.
\newblock Hyperstyle: Stylegan inversion with hypernetworks for real image editing.
\newblock In \emph{Proceedings of the IEEE/CVF conference on computer Vision and pattern recognition}, pp.\  18511--18521, 2022.

\bibitem[Bao et~al.(2023)Bao, Li, Huo, Ding, Liang, Li, and Gao]{bao2023and}
Yanqi Bao, Yuxin Li, Jing Huo, Tianyu Ding, Xinyue Liang, Wenbin Li, and Yang Gao.
\newblock Where and how: Mitigating confusion in neural radiance fields from sparse inputs.
\newblock \emph{arXiv preprint arXiv:2308.02908}, 2023.

\bibitem[Barron et~al.(2021)Barron, Mildenhall, Tancik, Hedman, Martin-Brualla, and Srinivasan]{barron2021mip}
Jonathan~T Barron, Ben Mildenhall, Matthew Tancik, Peter Hedman, Ricardo Martin-Brualla, and Pratul~P Srinivasan.
\newblock Mip-nerf: A multiscale representation for anti-aliasing neural radiance fields.
\newblock In \emph{Proceedings of the IEEE/CVF International Conference on Computer Vision}, pp.\  5855--5864, 2021.

\bibitem[Barron et~al.(2022)Barron, Mildenhall, Verbin, Srinivasan, and Hedman]{barron2022mip}
Jonathan~T Barron, Ben Mildenhall, Dor Verbin, Pratul~P Srinivasan, and Peter Hedman.
\newblock Mip-nerf 360: Unbounded anti-aliased neural radiance fields.
\newblock In \emph{Proceedings of the IEEE/CVF Conference on Computer Vision and Pattern Recognition}, pp.\  5470--5479, 2022.

\bibitem[Chan et~al.(2007)Chan, Shum, and Ng]{chan2007image}
SC~Chan, Heung-Yeung Shum, and King-To Ng.
\newblock Image-based rendering and synthesis.
\newblock \emph{IEEE Signal Processing Magazine}, 24\penalty0 (6):\penalty0 22--33, 2007.

\bibitem[Chauhan et~al.(2023)Chauhan, Zhou, Lu, Molaei, and Clifton]{chauhan2023brief}
Vinod~Kumar Chauhan, Jiandong Zhou, Ping Lu, Soheila Molaei, and David~A Clifton.
\newblock A brief review of hypernetworks in deep learning.
\newblock \emph{arXiv preprint arXiv:2306.06955}, 2023.

\bibitem[Chen et~al.(2021)Chen, Xu, Zhao, Zhang, Xiang, Yu, and Su]{chen2021mvsnerf}
Anpei Chen, Zexiang Xu, Fuqiang Zhao, Xiaoshuai Zhang, Fanbo Xiang, Jingyi Yu, and Hao Su.
\newblock Mvsnerf: Fast generalizable radiance field reconstruction from multi-view stereo.
\newblock In \emph{Proceedings of the IEEE/CVF International Conference on Computer Vision}, pp.\  14124--14133, 2021.

\bibitem[Chen et~al.(2022)Chen, Xu, Geiger, Yu, and Su]{chen2022tensorf}
Anpei Chen, Zexiang Xu, Andreas Geiger, Jingyi Yu, and Hao Su.
\newblock Tensorf: Tensorial radiance fields.
\newblock In \emph{European Conference on Computer Vision}, pp.\  333--350. Springer, 2022.

\bibitem[Chen et~al.(2019)Chen, Rohrbach, Yan, Shuicheng, Feng, and Kalantidis]{chen2019graph}
Yunpeng Chen, Marcus Rohrbach, Zhicheng Yan, Yan Shuicheng, Jiashi Feng, and Yannis Kalantidis.
\newblock Graph-based global reasoning networks.
\newblock In \emph{Proceedings of the IEEE/CVF Conference on Computer Vision and Pattern Recognition}, pp.\  433--442, 2019.

\bibitem[Chiang et~al.(2022)Chiang, Tsai, Tseng, Lai, and Chiu]{chiang2022stylizing}
Pei-Ze Chiang, Meng-Shiun Tsai, Hung-Yu Tseng, Wei-Sheng Lai, and Wei-Chen Chiu.
\newblock Stylizing 3d scene via implicit representation and hypernetwork.
\newblock In \emph{Proceedings of the IEEE/CVF Winter Conference on Applications of Computer Vision}, pp.\  1475--1484, 2022.

\bibitem[Debevec et~al.(1996)Debevec, Taylor, and Malik]{debevec1996modeling}
Paul~E Debevec, Camillo~J Taylor, and Jitendra Malik.
\newblock Modeling and rendering architecture from photographs: A hybrid geometry-and image-based approach.
\newblock In \emph{Proceedings of the 23rd annual conference on Computer graphics and interactive techniques}, pp.\  11--20, 1996.

\bibitem[Dupont et~al.(2020)Dupont, Martin, Colburn, Sankar, Susskind, and Shan]{dupont2020equivariant}
Emilien Dupont, Miguel~Bautista Martin, Alex Colburn, Aditya Sankar, Josh Susskind, and Qi~Shan.
\newblock Equivariant neural rendering.
\newblock In \emph{International Conference on Machine Learning}, pp.\  2761--2770. PMLR, 2020.

\bibitem[Genova et~al.(2020)Genova, Cole, Sud, Sarna, and Funkhouser]{genova2020local}
Kyle Genova, Forrester Cole, Avneesh Sud, Aaron Sarna, and Thomas Funkhouser.
\newblock Local deep implicit functions for 3d shape.
\newblock In \emph{Proceedings of the IEEE/CVF Conference on Computer Vision and Pattern Recognition}, pp.\  4857--4866, 2020.

\bibitem[Ha et~al.(2016)Ha, Dai, and Le]{ha2016hypernetworks}
David Ha, Andrew Dai, and Quoc~V Le.
\newblock Hypernetworks.
\newblock \emph{arXiv preprint arXiv:1609.09106}, 2016.

\bibitem[He et~al.(2016)He, Zhang, Ren, and Sun]{he2016deep}
Kaiming He, Xiangyu Zhang, Shaoqing Ren, and Jian Sun.
\newblock Deep residual learning for image recognition.
\newblock In \emph{Proceedings of the IEEE conference on computer vision and pattern recognition}, pp.\  770--778, 2016.

\bibitem[Huang et~al.(2023)Huang, Zhang, Feng, Li, Wang, and Wang]{huang2023local}
Xin Huang, Qi~Zhang, Ying Feng, Xiaoyu Li, Xuan Wang, and Qing Wang.
\newblock Local implicit ray function for generalizable radiance field representation.
\newblock In \emph{Proceedings of the IEEE/CVF Conference on Computer Vision and Pattern Recognition}, pp.\  97--107, 2023.

\bibitem[Iizuka et~al.(2017)Iizuka, Simo-Serra, and Ishikawa]{iizuka2017globally}
Satoshi Iizuka, Edgar Simo-Serra, and Hiroshi Ishikawa.
\newblock Globally and locally consistent image completion.
\newblock \emph{ACM Transactions on Graphics (ToG)}, 36\penalty0 (4):\penalty0 1--14, 2017.

\bibitem[Isaac-Medina et~al.(2023)Isaac-Medina, Willcocks, and Breckon]{isaac2023exact}
Brian~KS Isaac-Medina, Chris~G Willcocks, and Toby~P Breckon.
\newblock Exact-nerf: An exploration of a precise volumetric parameterization for neural radiance fields.
\newblock In \emph{Proceedings of the IEEE/CVF Conference on Computer Vision and Pattern Recognition}, pp.\  66--75, 2023.

\bibitem[Jain et~al.(2023)Jain, Kumar, and Van~Gool]{jain2023enhanced}
Nishant Jain, Suryansh Kumar, and Luc Van~Gool.
\newblock Enhanced stable view synthesis.
\newblock In \emph{Proceedings of the IEEE/CVF Conference on Computer Vision and Pattern Recognition}, pp.\  13208--13217, 2023.

\bibitem[Jiang et~al.(2020)Jiang, Sud, Makadia, Huang, Nie{\ss}ner, Funkhouser, et~al.]{jiang2020local}
Chiyu Jiang, Avneesh Sud, Ameesh Makadia, Jingwei Huang, Matthias Nie{\ss}ner, Thomas Funkhouser, et~al.
\newblock Local implicit grid representations for 3d scenes.
\newblock In \emph{Proceedings of the IEEE/CVF Conference on Computer Vision and Pattern Recognition}, pp.\  6001--6010, 2020.

\bibitem[Johari et~al.(2022)Johari, Lepoittevin, and Fleuret]{johari2022geonerf}
Mohammad~Mahdi Johari, Yann Lepoittevin, and Fran{\c{c}}ois Fleuret.
\newblock Geonerf: Generalizing nerf with geometry priors.
\newblock In \emph{Proceedings of the IEEE/CVF Conference on Computer Vision and Pattern Recognition}, pp.\  18365--18375, 2022.

\bibitem[Kania et~al.(2023)Kania, Kasymov, Zi{\k{e}}ba, and Spurek]{kania2023hypernerfgan}
Adam Kania, Artur Kasymov, Maciej Zi{\k{e}}ba, and Przemys{\l}aw Spurek.
\newblock Hypernerfgan: Hypernetwork approach to 3d nerf gan.
\newblock \emph{arXiv preprint arXiv:2301.11631}, 2023.

\bibitem[Kingma \& Welling(2013)Kingma and Welling]{kingma2013auto}
Diederik~P Kingma and Max Welling.
\newblock Auto-encoding variational bayes.
\newblock \emph{arXiv preprint arXiv:1312.6114}, 2013.

\bibitem[Knapitsch et~al.(2017)Knapitsch, Park, Zhou, and Koltun]{knapitsch2017tanks}
Arno Knapitsch, Jaesik Park, Qian-Yi Zhou, and Vladlen Koltun.
\newblock Tanks and temples: Benchmarking large-scale scene reconstruction.
\newblock \emph{ACM Transactions on Graphics (ToG)}, 36\penalty0 (4):\penalty0 1--13, 2017.

\bibitem[Kopf et~al.(2013)Kopf, Langguth, Scharstein, Szeliski, and Goesele]{kopf2013image}
Johannes Kopf, Fabian Langguth, Daniel Scharstein, Richard Szeliski, and Michael Goesele.
\newblock Image-based rendering in the gradient domain.
\newblock \emph{ACM Transactions on Graphics (TOG)}, 32\penalty0 (6):\penalty0 1--9, 2013.

\bibitem[Kulh{\'a}nek et~al.(2022)Kulh{\'a}nek, Derner, Sattler, and Babu{\v{s}}ka]{kulhanek2022viewformer}
Jon{\'a}{\v{s}} Kulh{\'a}nek, Erik Derner, Torsten Sattler, and Robert Babu{\v{s}}ka.
\newblock Viewformer: Nerf-free neural rendering from few images using transformers.
\newblock In \emph{European Conference on Computer Vision}, pp.\  198--216. Springer, 2022.

\bibitem[Li et~al.(2020)Li, Gu, Zhang, Van~Gool, and Timofte]{li2020dhp}
Yawei Li, Shuhang Gu, Kai Zhang, Luc Van~Gool, and Radu Timofte.
\newblock Dhp: Differentiable meta pruning via hypernetworks.
\newblock In \emph{Computer Vision--ECCV 2020: 16th European Conference, Glasgow, UK, August 23--28, 2020, Proceedings, Part VIII 16}, pp.\  608--624. Springer, 2020.

\bibitem[Lin \& Shum(2004)Lin and Shum]{lin2004geometric}
Zhouchen Lin and Heung-Yeung Shum.
\newblock A geometric analysis of light field rendering.
\newblock \emph{International Journal of Computer Vision}, 58:\penalty0 121--138, 2004.

\bibitem[Liu et~al.(2019)Liu, Saito, Chen, and Li]{liu2019learning}
Shichen Liu, Shunsuke Saito, Weikai Chen, and Hao Li.
\newblock Learning to infer implicit surfaces without 3d supervision.
\newblock \emph{Advances in Neural Information Processing Systems}, 32, 2019.

\bibitem[Liu et~al.(2022)Liu, Peng, Liu, Wang, Wang, Theobalt, Zhou, and Wang]{liu2022neural}
Yuan Liu, Sida Peng, Lingjie Liu, Qianqian Wang, Peng Wang, Christian Theobalt, Xiaowei Zhou, and Wenping Wang.
\newblock Neural rays for occlusion-aware image-based rendering.
\newblock In \emph{Proceedings of the IEEE/CVF Conference on Computer Vision and Pattern Recognition}, pp.\  7824--7833, 2022.

\bibitem[Liu et~al.(2023)Liu, Feng, Black, Nowrouzezahrai, Paull, and Liu]{liu2023meshdiffusion}
Zhen Liu, Yao Feng, Michael~J Black, Derek Nowrouzezahrai, Liam Paull, and Weiyang Liu.
\newblock Meshdiffusion: Score-based generative 3d mesh modeling.
\newblock \emph{arXiv preprint arXiv:2303.08133}, 2023.

\bibitem[Mescheder et~al.(2019)Mescheder, Oechsle, Niemeyer, Nowozin, and Geiger]{mescheder2019occupancy}
Lars Mescheder, Michael Oechsle, Michael Niemeyer, Sebastian Nowozin, and Andreas Geiger.
\newblock Occupancy networks: Learning 3d reconstruction in function space.
\newblock In \emph{Proceedings of the IEEE/CVF conference on computer vision and pattern recognition}, pp.\  4460--4470, 2019.

\bibitem[Mildenhall et~al.(2021)Mildenhall, Srinivasan, Tancik, Barron, Ramamoorthi, and Ng]{mildenhall2021nerf}
Ben Mildenhall, Pratul~P Srinivasan, Matthew Tancik, Jonathan~T Barron, Ravi Ramamoorthi, and Ren Ng.
\newblock Nerf: Representing scenes as neural radiance fields for view synthesis.
\newblock \emph{Communications of the ACM}, 65\penalty0 (1):\penalty0 99--106, 2021.

\bibitem[Niemeyer et~al.(2022)Niemeyer, Barron, Mildenhall, Sajjadi, Geiger, and Radwan]{niemeyer2022regnerf}
Michael Niemeyer, Jonathan~T Barron, Ben Mildenhall, Mehdi~SM Sajjadi, Andreas Geiger, and Noha Radwan.
\newblock Regnerf: Regularizing neural radiance fields for view synthesis from sparse inputs.
\newblock In \emph{Proceedings of the IEEE/CVF Conference on Computer Vision and Pattern Recognition}, pp.\  5480--5490, 2022.

\bibitem[Peng et~al.(2023)Peng, Yan, Shuai, Bao, and Zhou]{peng2023representing}
Sida Peng, Yunzhi Yan, Qing Shuai, Hujun Bao, and Xiaowei Zhou.
\newblock Representing volumetric videos as dynamic mlp maps.
\newblock In \emph{Proceedings of the IEEE/CVF Conference on Computer Vision and Pattern Recognition}, pp.\  4252--4262, 2023.

\bibitem[Peng et~al.(2020)Peng, Niemeyer, Mescheder, Pollefeys, and Geiger]{peng2020convolutional}
Songyou Peng, Michael Niemeyer, Lars Mescheder, Marc Pollefeys, and Andreas Geiger.
\newblock Convolutional occupancy networks.
\newblock In \emph{Computer Vision--ECCV 2020: 16th European Conference, Glasgow, UK, August 23--28, 2020, Proceedings, Part III 16}, pp.\  523--540. Springer, 2020.

\bibitem[Perez et~al.(2018)Perez, Strub, De~Vries, Dumoulin, and Courville]{perez2018film}
Ethan Perez, Florian Strub, Harm De~Vries, Vincent Dumoulin, and Aaron Courville.
\newblock Film: Visual reasoning with a general conditioning layer.
\newblock In \emph{Proceedings of the AAAI conference on artificial intelligence}, volume~32, 2018.

\bibitem[Poole et~al.(2022)Poole, Jain, Barron, and Mildenhall]{poole2022dreamfusion}
Ben Poole, Ajay Jain, Jonathan~T Barron, and Ben Mildenhall.
\newblock Dreamfusion: Text-to-3d using 2d diffusion.
\newblock \emph{arXiv preprint arXiv:2209.14988}, 2022.

\bibitem[Riegler \& Koltun(2020)Riegler and Koltun]{riegler2020free}
Gernot Riegler and Vladlen Koltun.
\newblock Free view synthesis.
\newblock In \emph{Computer Vision--ECCV 2020: 16th European Conference, Glasgow, UK, August 23--28, 2020, Proceedings, Part XIX 16}, pp.\  623--640. Springer, 2020.

\bibitem[Ronneberger et~al.(2015)Ronneberger, Fischer, and Brox]{ronneberger2015u}
Olaf Ronneberger, Philipp Fischer, and Thomas Brox.
\newblock U-net: Convolutional networks for biomedical image segmentation.
\newblock In \emph{Medical Image Computing and Computer-Assisted Intervention--MICCAI 2015: 18th International Conference, Munich, Germany, October 5-9, 2015, Proceedings, Part III 18}, pp.\  234--241. Springer, 2015.

\bibitem[Sinha et~al.(2009)Sinha, Steedly, and Szeliski]{sinha2009piecewise}
Sudipta Sinha, Drew Steedly, and Rick Szeliski.
\newblock Piecewise planar stereo for image-based rendering.
\newblock In \emph{2009 International Conference on Computer Vision}, pp.\  1881--1888, 2009.

\bibitem[Sitzmann et~al.(2019)Sitzmann, Zollh{\"o}fer, and Wetzstein]{sitzmann2019scene}
Vincent Sitzmann, Michael Zollh{\"o}fer, and Gordon Wetzstein.
\newblock Scene representation networks: Continuous 3d-structure-aware neural scene representations.
\newblock \emph{Advances in Neural Information Processing Systems}, 32, 2019.

\bibitem[Sitzmann et~al.(2020)Sitzmann, Martel, Bergman, Lindell, and Wetzstein]{sitzmann2020implicit}
Vincent Sitzmann, Julien Martel, Alexander Bergman, David Lindell, and Gordon Wetzstein.
\newblock Implicit neural representations with periodic activation functions.
\newblock \emph{Advances in neural information processing systems}, 33:\penalty0 7462--7473, 2020.

\bibitem[Spurek et~al.(2020{\natexlab{a}})Spurek, Winczowski, Tabor, Zamorski, Zi{\k{e}}ba, and Trzci{\'n}ski]{spurek2020hypernetwork}
Przemys{\l}aw Spurek, Sebastian Winczowski, Jacek Tabor, Maciej Zamorski, Maciej Zi{\k{e}}ba, and Tomasz Trzci{\'n}ski.
\newblock Hypernetwork approach to generating point clouds.
\newblock \emph{arXiv preprint arXiv:2003.00802}, 2020{\natexlab{a}}.

\bibitem[Spurek et~al.(2020{\natexlab{b}})Spurek, Zi{\k{e}}ba, Tabor, and Trzci{\'n}ski]{spurek2020hyperflow}
Przemys{\l}aw Spurek, Maciej Zi{\k{e}}ba, Jacek Tabor, and Tomasz Trzci{\'n}ski.
\newblock Hyperflow: Representing 3d objects as surfaces.
\newblock \emph{arXiv preprint arXiv:2006.08710}, 2020{\natexlab{b}}.

\bibitem[Suhail et~al.(2022)Suhail, Esteves, Sigal, and Makadia]{suhail2022generalizable}
Mohammed Suhail, Carlos Esteves, Leonid Sigal, and Ameesh Makadia.
\newblock Generalizable patch-based neural rendering.
\newblock In \emph{European Conference on Computer Vision}, pp.\  156--174. Springer, 2022.

\bibitem[Von~Oswald et~al.(2019)Von~Oswald, Henning, Grewe, and Sacramento]{von2019continual}
Johannes Von~Oswald, Christian Henning, Benjamin~F Grewe, and Jo{\~a}o Sacramento.
\newblock Continual learning with hypernetworks.
\newblock \emph{arXiv preprint arXiv:1906.00695}, 2019.

\bibitem[Wang et~al.(2018)Wang, Samari, and Siddiqi]{wang2018local}
Chu Wang, Babak Samari, and Kaleem Siddiqi.
\newblock Local spectral graph convolution for point set feature learning.
\newblock In \emph{Proceedings of the European conference on computer vision (ECCV)}, pp.\  52--66, 2018.

\bibitem[Wang et~al.(2022)Wang, Chen, Chen, Venugopalan, Wang, et~al.]{wang2022attention}
Peihao Wang, Xuxi Chen, Tianlong Chen, Subhashini Venugopalan, Zhangyang Wang, et~al.
\newblock Is attention all nerf needs?
\newblock \emph{arXiv preprint arXiv:2207.13298}, 2022.

\bibitem[Wang et~al.(2021)Wang, Wang, Genova, Srinivasan, Zhou, Barron, Martin-Brualla, Snavely, and Funkhouser]{wang2021ibrnet}
Qianqian Wang, Zhicheng Wang, Kyle Genova, Pratul~P Srinivasan, Howard Zhou, Jonathan~T Barron, Ricardo Martin-Brualla, Noah Snavely, and Thomas Funkhouser.
\newblock Ibrnet: Learning multi-view image-based rendering.
\newblock In \emph{Proceedings of the IEEE/CVF Conference on Computer Vision and Pattern Recognition}, pp.\  4690--4699, 2021.

\bibitem[Wang et~al.(2004)Wang, Bovik, Sheikh, and Simoncelli]{wang2004image}
Zhou Wang, Alan~C Bovik, Hamid~R Sheikh, and Eero~P Simoncelli.
\newblock Image quality assessment: from error visibility to structural similarity.
\newblock \emph{IEEE transactions on image processing}, 13\penalty0 (4):\penalty0 600--612, 2004.

\bibitem[Xu et~al.(2023)Xu, Zhan, Zhang, Yu, Zhang, Theobalt, Shao, and Lu]{xu2023wavenerf}
Muyu Xu, Fangneng Zhan, Jiahui Zhang, Yingchen Yu, Xiaoqin Zhang, Christian Theobalt, Ling Shao, and Shijian Lu.
\newblock Wavenerf: Wavelet-based generalizable neural radiance fields.
\newblock \emph{arXiv preprint arXiv:2308.04826}, 2023.

\bibitem[Yang et~al.(2023)Yang, Pavone, and Wang]{yang2023freenerf}
Jiawei Yang, Marco Pavone, and Yue Wang.
\newblock Freenerf: Improving few-shot neural rendering with free frequency regularization.
\newblock In \emph{Proceedings of the IEEE/CVF Conference on Computer Vision and Pattern Recognition}, pp.\  8254--8263, 2023.

\bibitem[Yang et~al.(2022)Yang, Li, Song, Zhao, Tao, Zhou, Liang, and Yang]{yang2022dynamic}
Lingfeng Yang, Xiang Li, Renjie Song, Borui Zhao, Juntian Tao, Shihao Zhou, Jiajun Liang, and Jian Yang.
\newblock Dynamic mlp for fine-grained image classification by leveraging geographical and temporal information.
\newblock In \emph{Proceedings of the IEEE/CVF Conference on Computer Vision and Pattern Recognition}, pp.\  10945--10954, 2022.

\bibitem[Yao et~al.(2018)Yao, Luo, Li, Fang, and Quan]{yao2018mvsnet}
Yao Yao, Zixin Luo, Shiwei Li, Tian Fang, and Long Quan.
\newblock Mvsnet: Depth inference for unstructured multi-view stereo.
\newblock In \emph{Proceedings of the European conference on computer vision (ECCV)}, pp.\  767--783, 2018.

\bibitem[Yu et~al.(2021)Yu, Ye, Tancik, and Kanazawa]{yu2021pixelnerf}
Alex Yu, Vickie Ye, Matthew Tancik, and Angjoo Kanazawa.
\newblock pixelnerf: Neural radiance fields from one or few images.
\newblock In \emph{Proceedings of the IEEE/CVF Conference on Computer Vision and Pattern Recognition}, pp.\  4578--4587, 2021.

\bibitem[Zamorski et~al.(2020)Zamorski, Zi{\k{e}}ba, Klukowski, Nowak, Kurach, Stokowiec, and Trzci{\'n}ski]{zamorski2020adversarial}
Maciej Zamorski, Maciej Zi{\k{e}}ba, Piotr Klukowski, Rafa{\l} Nowak, Karol Kurach, Wojciech Stokowiec, and Tomasz Trzci{\'n}ski.
\newblock Adversarial autoencoders for compact representations of 3d point clouds.
\newblock \emph{Computer Vision and Image Understanding}, 193:\penalty0 102921, 2020.

\bibitem[Zhang et~al.(2020)Zhang, Riegler, Snavely, and Koltun]{zhang2020nerf++}
Kai Zhang, Gernot Riegler, Noah Snavely, and Vladlen Koltun.
\newblock Nerf++: Analyzing and improving neural radiance fields.
\newblock \emph{arXiv preprint arXiv:2010.07492}, 2020.

\bibitem[Zhang et~al.(2018)Zhang, Isola, Efros, Shechtman, and Wang]{zhang2018unreasonable}
Richard Zhang, Phillip Isola, Alexei~A Efros, Eli Shechtman, and Oliver Wang.
\newblock The unreasonable effectiveness of deep features as a perceptual metric.
\newblock In \emph{Proceedings of the IEEE conference on computer vision and pattern recognition}, pp.\  586--595, 2018.

\bibitem[Zhang et~al.(2023)Zhang, Liu, Han, Pan, Guo, and Yao]{zhang2023transforming}
Zicheng Zhang, Yinglu Liu, Congying Han, Yingwei Pan, Tiande Guo, and Ting Yao.
\newblock Transforming radiance field with lipschitz network for photorealistic 3d scene stylization.
\newblock In \emph{Proceedings of the IEEE/CVF Conference on Computer Vision and Pattern Recognition}, pp.\  20712--20721, 2023.

\bibitem[Zhou et~al.(2021)Zhou, Du, and Wu]{zhou20213d}
Linqi Zhou, Yilun Du, and Jiajun Wu.
\newblock 3d shape generation and completion through point-voxel diffusion.
\newblock In \emph{Proceedings of the IEEE/CVF International Conference on Computer Vision}, pp.\  5826--5835, 2021.

\bibitem[Zhu et~al.(2023)Zhu, Ding, Chen, Zharkov, Nevatia, and Liang]{zhu2023caesarnerf}
Haidong Zhu, Tianyu Ding, Tianyi Chen, Ilya Zharkov, Ram Nevatia, and Luming Liang.
\newblock Caesarnerf: Calibrated semantic representation for few-shot generalizable neural rendering.
\newblock \emph{arXiv preprint arXiv:2311.15510}, 2023.

\bibitem[Zimny et~al.(2022)Zimny, Trzci{\'n}ski, and Spurek]{zimny2022points2nerf}
Dominik Zimny, T~Trzci{\'n}ski, and Przemys{\l}aw Spurek.
\newblock Points2nerf: Generating neural radiance fields from 3d point cloud.
\newblock \emph{arXiv preprint arXiv:2206.01290}, 2022.

\bibitem[Zimny et~al.(2023)Zimny, Tabor, Zi{\k{e}}ba, and Spurek]{zimny2023multiplanenerf}
Dominik Zimny, Jacek Tabor, Maciej Zi{\k{e}}ba, and Przemys{\l}aw Spurek.
\newblock Multiplanenerf: Neural radiance field with non-trainable representation.
\newblock \emph{arXiv preprint arXiv:2305.10579}, 2023.

\end{thebibliography}
\bibliographystyle{iclr2024_conference}

\appendix
\section{Experimental Details}\label{A}
 During the training process, the feature extraction and InsertNeRF-systems are trained with different learning rates in an end-to-end manner, and we apply the Adam to optimize the entire network with an exponentially decaying learning rate. In our experiments, $\lambda_1$ is set as 0.1 and $\lambda_2$ is set as 1. During the cross-scene evaluation phase, we individually test the average PSNR, SSIM, and LPIPS metrics of all testing-view renderings for each scene and report the average values across all scenes, as shown in~\Cref{tab1}. For the sake of fair experimental comparison and efficiency with~\citep{wang2022attention}, we set the rendering stride size to 2 across all experiments. Empirical evidence demonstrates that this choice influences the evaluation of metrics, as shown in~\Cref{tab8}. Our model is implemented using PyTorch 1.11.0 and all experiments are conducted on Nvidia RTX 3090 GPUs with CUDA 11.4. 

\textbf{Metrics.} We calculate the Peak Signal to-Noise Ratio (PSNR) and Structural SIMilarity (SSIM) ~\citep{wang2004image} to evaluate rendering quality for target novel viewpoints. Additionally, the Learned Perceptual Image Patch Similarity (LPIPS) ~\citep{zhang2018unreasonable} is adopted as a perceptual metric. For all our experiments, we report the average of the PSNR, SSIM and LPIPS under different testing views in multiple scenes to verify the generalization. It is noteworthy that, similar to the majority of GNeRF research~\citep{liu2022neural,wang2022attention}, our focus lies on foreground-centric metrics for both the NeRF Synthetic and DTU datasets during the evaluation process.

\subsection{Evaluation Datasets.}
\textbf{NeRF Synthetic.} The dataset consists of 8 synthetic objects with viewpoints uniformly sampled on the upper hemisphere. Each scene comprises 200 test images, wherein a sampling strategy with an interval of 8 images is employed during the evaluation phase.\\
\textbf{Local Light Field Fusion (LLFF).} The dataset consists of 8 complex real-world scenes. Each scene includes real images, 1/8 of which are used as the evaluation dataset.\\
\textbf{DTU Dataset.} The dataset consists of 128 object-scenes, which is a classic dataset of MVS. In the experiments, we select 4 scenes (birds, tools, bricks and snowman) for evaluation~\citep{liu2022neural}.

\subsection{Experimental Details for InsertNeRF}
For all our experiments, we maintain the experimental protocols of NeuRay~\citep{liu2022neural} and GNT~\citep{wang2022attention}. In Setting I, we also employ depth maps~\citep{liu2022neural} as priors to assist the Hypernet modules to generate adaptive weights. Following~\citep{liu2022neural}, we randomly sample 2048 rays from Target-Reference pairs, and it trains for a total of 600,000 steps. In Setting II~\citep{wang2022attention}, we randomly sample 512 rays from Target-Reference pairs, and it trains for a total of 400,000 steps without any priors. In order to enhance training and inference efficiency, we sample $K=64$ points along each ray and simplify the volume rendering process in our paradigm.
\subsection{Experimental Details for Insert-mip-NeRF}
Insert-mip-NeRF substitutes point-samplings with a sequence of conical-frustums and introduces the integrated positional encoding into the InsertNeRF framework. In contrast to InsertNeRF, we sample $K=65$ positions for integrated positional encoding on conical-frustums. Throughout all experiments, we employ the multi-scale variants of NeRF Synthetic, such as Full, 1/2, 1/4 and 1/8 Resolutions, to simulate multi-resolution scenes for training and evaluation as~\citep{barron2021mip}. The remaining experimental configurations follow those of mipNeRF and our Setting II.
\subsection{Experimental Details for Insert-NeRF++}
Insert-NeRF++ divides the InsertNeRF's scene representations into foreground and background components and combines both for finally rendering~\citep{zhang2020nerf++}. In Insert-NeRF++, we avoid using inverted sphere parametrization and chose standard parameterization across foreground-background spatial ranges, which possesses the capability to offer precise spatial positions for projection process $\Pi_n(\bm x)$. Following~\citep{zhang2020nerf++}, we employ the Tanks and Temples dataset~\citep{knapitsch2017tanks}, a real-world unbound dataset captured with hand-held cameras, for training and evaluation under cross-scenes setting. During the evaluation process, we separately report the renderings for the foreground and background as~\citep{zhang2020nerf++}. The remaining experimental configurations also follow those of vanilla NeRF++ and our Setting II.
\section{Related Works}\label{B}
\subsection{Image-based Rendering}
Image-based rendering (IBR)~\citep{chan2007image} is a classic technique that aims to generate novel views within a specific scene by warping and integrating pixel information from reference-view images. To ensure spatial consistency, most existing works simplify this challenge by resorting to estimated explicit geometry or depth maps ~\citep{riegler2020free, jain2023enhanced}. In order to obtain proxy geometry, Structure-from-Motion (SfM)~\citep{sinha2009piecewise} and MultiView Stereo (MVS)~\citep{yao2018mvsnet} have recently attracted the attention of researchers in the field of IBR. However, hints from explicit geometry without 3D supervision are unstable in most scenarios. Recently light field rendering~\citep{lin2004geometric} has become one of the alternatives to explicit representations, which considers the lighting and reflection properties of the scene to ensure visually plausible rendering. Moreover, some works~\citep{kopf2013image, chen2021mvsnerf} focus on aggregating information from multiple reference views, which exploits the relationships between references and also implicitly solves the occlusion. As opposed to explicit geometry, these methods rely on constructing implicit representations to enable reasoning about novel views~\citep{liu2019learning}. Different from the above explicit or implicit scene-customized representations for IBR, our method can be used in a large number of scenes simultaneously without retraining.

\subsection{Neural Scene Representation}
Representing the geometry and appearance of a scenes with neural networks has been considered an alternative to 3D scene representations in recent years~\citep{mescheder2019occupancy, peng2020convolutional}. Existing works demonstrate the potential of Multi-Layer Perceptrons (MLPs) in implicit representations, which activate spatial features by continuous functions~\citep{genova2020local, jiang2020local}. Neural radiance fields (NeRF)~\citep{mildenhall2021nerf} apply such functions for coordinate-based representations, which use high-dimensional interpolation to produce photorealistic renderings of target views. On its basis, mip-NeRF~\citep{barron2021mip} replaces rays with casting cones during volume rendering, changing the input of NeRF from points to cone frustums, and introduces an integrated positional encoding for multi-resolutions images. Subsequently, NeRF++~\citep{zhang2020nerf++} and
mip-NeRF 360~\citep{barron2022mip} further improve NeRF and mip-NeRF to adapt to distant targets under unbounded scenes. To further enhance representation efficiency, \citep{chen2022tensorf} and \citep{zimny2023multiplanenerf} proposed efficient representation methods to replace a large number of neural network parameters. Although derivative works has been a surge, similar to most IBR works, NeRF must also be trained for each novel scene, which is time-consuming in practice.

\subsection{Generative Models}
With the development of generative models, 3D generative models have been widely discussed, enabling the direct construction of 3D representations such as
point clouds~\citep{zamorski2020adversarial}, surfaces~\citep{spurek2020hyperflow}, voxels~\citep{zhou20213d} and NeRF~\citep{poole2022dreamfusion}. A significant amount of works have leveraged techniques from image generative models and applied them to 3D generation, including GAN ~\citep{kania2023hypernerfgan} and diffusion model~\citep{liu2023meshdiffusion}. In this work, we focus on some 3D generative models with HyperNetwork. \citep{spurek2020hypernetwork} is an early work that builds variable size representations of point clouds with HyperNetwork. Then, HyperFlow~\citep{spurek2020hyperflow} uses a hypernetwork to model 3D objects as families of surfaces and Points2NeRF~\citep{zimny2022points2nerf} utilizes a HyperNetwork to generate NeRF from a 3D point cloud. Additionally, in recent years, there have been some NeRF works that focus on this technique, they directly incorporate HyperNetwork into NeRF, as described in Section 2.2. However, in the generalizable NeRF task, such idea is suboptimal, overlooking the characteristic of different attributes, such as volume density and color. Furthermore, they struggle to capture the relationship between the inputs (reference images) in the target's sampling process. Therefore, this paper proposes two types of HyperNet module structures for $\mathcal{F}_{geo}$ and $\mathcal{F}_{app}$ and Sampling-aware Filter separately to mitigate the aforementioned two issues.

\section{Implementation Details}
\subsection{HyperNet Module Architecture}\label{C1}
\begin{figure}[h]
\begin{center}
\subfigure[HyperNet module in $\mathcal{F}_{geo}$]{
\label{fig6a}
\includegraphics[width=6.8cm,height = 2.2cm]{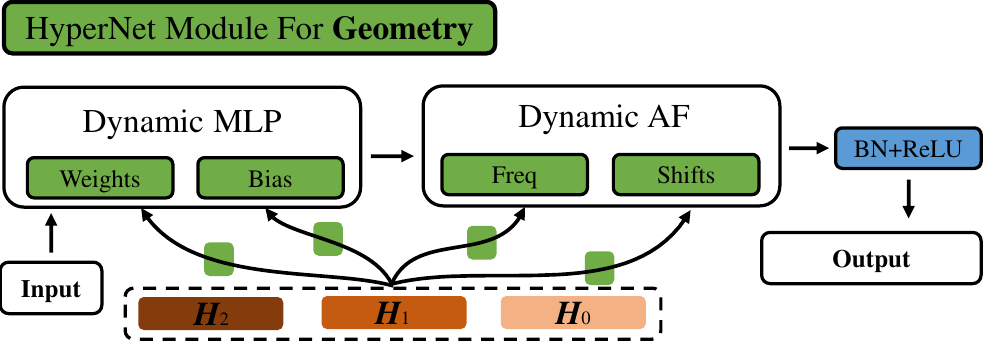}}\subfigure[HyperNet module in $\mathcal{F}_{app}$]{
\label{fig6b}
\includegraphics[width=6.8cm,height = 2.2cm]{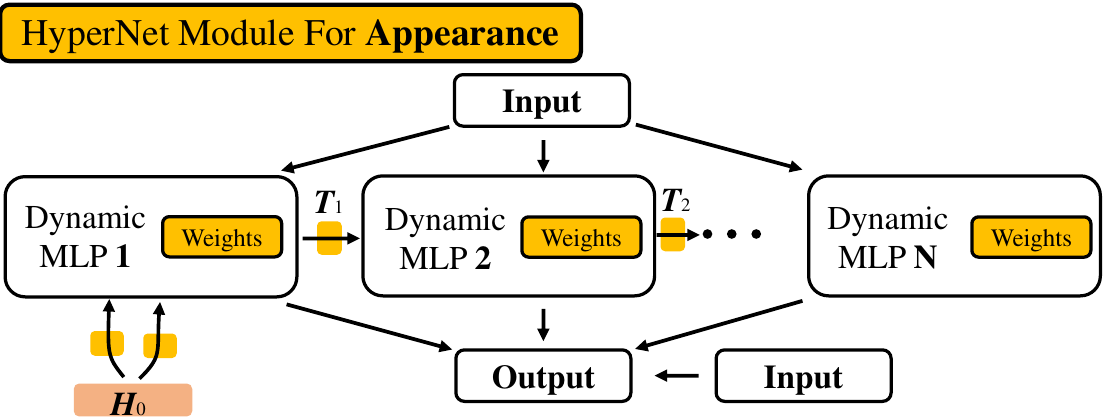}}
\caption{Detailed HyperNet module of $\mathcal{F}_{geo}$ and $\mathcal{F}_{app}$}
\end{center}
\end{figure}

Diverging from the conventional HyperNetworks~\citep{ha2016hypernetworks}, the direct prediction of distinct attributes, such as emitted color and volume density, within the NeRF framework often proves suboptimal. Drawing an analogy to the vanilla
NeRF~\citep{mildenhall2021nerf}, where different network depths were assigned for $\mathcal{F}_{geo}$ and $\mathcal{F}_{app}$, it is essential for InsertNeRF to discuss distinct HyperNet module structures for them. $\mathcal{F}_{geo}$ plays a pivotal role in the NeRF's geometric representations, which necessitates accurate inference and completion of the relationship between references based on global-local features. Based on this analysis, multi-scale features $\bm {H}_{l}$ are systematically introduced into the $\mathcal{F}_{geo}$'s HyperNet modules, as shown in~\Cref{fig6a}. After~\Cref{eq7}, $\bm F_{\text{output}}$ is subsequently fed into an original NeRF's MLP layer. Note that we also incorporate an additional ReLU and BatchNorm after MLP to ensure training stability for HyperNet modules. 

\begin{equation}
\bm {F}_\text{final} = \text{ReLU}(\text{BatchNorm}(\text{MLP}_{l}(\bm {F}_\text{output}
))) ,\label{eq10}
\end{equation}

Note that in $\mathcal{F}_{geo}$, as the network's depth increases, we leverage denser features for guiding weights prediction, i.e., global to dense. Additionally, when the number of MLP layers surpasses that of the feature layers, $\bm {H}_{0}$ will be recurrently utilized in the remaining MLP layers.

Unlike $\mathcal{F}_{geo}$, $\mathcal{F}_{app}$ exhibits a heightened focus on dense features~\citep{wang2022attention} and the smooth BRDF prior for surface reflectance~\citep{zhang2020nerf++}. As shown in~\Cref{fig6b}, $\mathcal{F}_{app}$'s HyperNet modules employ a parallel progressive generation paradigm and residual connection that respond to the desired smoothness. Specifically, given dense feature $\bm {H}_{0}$,
\begin{equation}
\bm {\tilde{F}}_\text{final} = \text{MLP}\left (\sum_{z=0}^{Z} \left ( \text{Weight}_{\bm{T}_{z}}\times \bm{F}_\text{input} \right ) + \bm{F}_\text{input}\right ), \quad \left\{\begin{matrix}
  \bm{T}_{z} = \text{Weight}_{\bm{T}_{z-1}}& z\ge 1\\
  \bm{T}_{z} = \bm {H}_{0} & z=0
\end{matrix}\right., \label{eq11}
\end{equation}
where $Z$ represents the number of parallel branches. Note that DFiLM and dynamic bias are not utilized in $\mathcal{F}_{app}$ for the smooth BRDF prior.

\subsection{Pseudocode}\label{C2}
In contrast to the scene-customized vanilla NeRF and its derivative works, GNeRF primarily concentrates on cross-scene rendering tasks without any retraining. As shown in~\Cref{fig1}, our HyperNet modules possess the capacity to instill generalizability into NeRF-like systems. Here, we delve further into the training processes of InsertNeRF-like systems. In Algorithm 1, due to $\bm\Omega_T^{\text{TrScene}}$'s adaptive response to the stochastic sampling of scenes by $\mathcal{D}_\text{TrScene}$ (that represents data of training scenes), InsertNeRF-like systems acquires inherent generalizability, where NeRF-like systems encompass diverse frameworks, including but not limited to mipNeRF, NeRF++, NeRF- -, and others. In the evaluation phase, for any given $\bm P_T$, we sample neighboring views $\left \{ \bm I_{n}, \bm P_n\right \}_{n=1}^N$ from $\mathcal{D}_\text{TeScene}$, rendering for $\bm I_T$ with the pretrained ${\bm \Theta}_\text{NeRF-like Systems}$ and ${\bm \Theta}_\text{HyperNet}$, as shown in Algorithm 2.

\begin{algorithm}
\caption{Training for InsertNeRF-like Systems}\label{algorithm1}
\KwData{Training Datasets $\mathcal{D}_\text{Train}$}
\KwResult{${\bm \Theta}_\text{NeRF-like Systems}$, ${\bm \Theta}_\text{HyperNet}$}
\While{$t\leq it$}
{$\textbf{Sample:} \; \bm r(t_i)   \leftarrow \left \{ \left \{ \bm I_{T}, \bm P_T\right \}, \left \{ \bm I_{n}, \bm P_n\right \}_{n=1}^N \right \} \leftarrow \mathcal{D}_\text{TrScene}\leftarrow \mathcal{D}_\text{Train}\;$\\
$\bm\Omega_T^{\text{TrScene}} \leftarrow \  \text{HyperNet} \left ( \left\{\mathcal{F}_{\text{view}}\left(\left\{\bm F_n\left( \Pi_n(\bm r(t_i)) \right)\right\}_{n=1}^N\right)\right\}_{i=1}^K\right )$\\
$\text{InsertNeRF-like Systems} \leftarrow \text{NeRF-like Systems} \left ( r(t_i), \bm\Omega_T^{\text{TrScene}} \right )$\\
${\bm \Theta}_\text{NeRF-like Systems}, {\bm \Theta}_\text{HyperNet} \leftarrow \mathcal{L}$\\
$t\leftarrow t+1$\;}
\end{algorithm}

\begin{algorithm} \SetKwInOut{Input}{Input} \SetKwInOut{Output}{Output}
\caption{Testing for InsertNeRF-like Systems}\label{algorithm2}
\Input{${\bm \Theta}_\text{NeRF-like Systems}$, ${\bm \Theta}_\text{HyperNet}$, $\mathcal{D}_\text{Test}$, and Random $\bm P_T$ in Testing Scenes.}
\Output {$\bm{I}_{T}$}
$\textbf{Sample:} \; \bm r(t_i)   \leftarrow \left \{\left \{ \left \{ \bm I_{n}, \bm P_n\right \}_{n=1}^N \leftarrow \mathcal{D}_\text{TeScene}\leftarrow \mathcal{D}_\text{Test}\right\}, \left \{\bm P_T\right \} \right \}$\\

$\;$$\;$$\;$$\;$$\bm\Omega_T^{\text{TeScene}} \leftarrow \  \text{HyperNet} \left ( \left\{\mathcal{F}_{\text{view}}\left(\left\{\bm F_n\left( \Pi_n(\bm r(t_i)) \right)\right\}_{n=1}^N\right)\right\}_{i=1}^K\right )$\\

$\;$$\;$$\;$$\;$$\bm{I}_{T} \leftarrow \text{InsertNeRF-like Systems} \left ( \bm{r}(t_i), \bm\Omega_T^{\text{TeScene}} \right )$\\
\end{algorithm}

\subsection{Graph Reasoning}
Graph-based methods have been the focus of extensive research recently and shown to be an efficient way of relation reasoning~\citep{wang2018local}. Following the spatial properties, we conceptualize all sampled points along a ray in a fully connected graph to find correlations between inter-samples and further update node features. Specifically, as shown in~\Cref{eq6}, InsertNeRF initially predicts a learnable adjacency matrix $\bm {A}_{l}$ to parameterize the edge weights between nodes, which models the relationships between sampled points. Subsequently, $\bm {W}^a_{l}$ is employed to update node states, mitigating the noise from epipolar geometric constrains. Furthermore, the identity matrix $\bm {I}$ is introduced to guide the learning process to pay more attention to the intrinsic characteristics of node features. An naive approach is to calculate the adjacency matrix based on the similarity between node features or Euclidean distance, and update node states, similar to existing graph convolution works. However, this is computationally expensive, especially for a large number of MLP blocks in InsertNeRF. Inspired by~\cite{chen2019graph}, $\bm {A}_{l}$ and $\bm {W}^a_{l}$ are replaced by two separate linear layers operating in different dimensions, while the identity matrix is represented as a residual connection,
\begin{equation}
\bm {H}_{l} =\text{Linear}\left ( \text{Linear}\left ( F_{\text{view}}\right )^T \right )^T + F_{\text{view}}.
\label{eq12}
\end{equation}

\begin{table}
    \centering
    \caption{Ablation studys for Multi-layer Dynamic-Static Aggregation Strategy with IBRNet~\citep{wang2021ibrnet}}
    \setlength{\tabcolsep}{1.5mm}
    \scalebox{1.0}{
    \begin{tabular}{l|ccc|ccc}
        \toprule[2pt]
            \multirow{2}{*}{Methods}& \multicolumn{3}{c|}{NeRF Synthetic} & \multicolumn{3}{c}{LLFF} \\
          & PSNR$\uparrow$ & SSIM$\uparrow$ & LPIPS$\downarrow$ & PSNR$\uparrow$ & SSIM$\uparrow$ & LPIPS$\downarrow$ \\
        \midrule
        InsertNeRF w/o Max &27.44  &0.930 &0.061  &25.52  &0.854 & 0.131 \\
        IBR-InsertNeRF &25.71  &0.909 &0.085 &25.00 &0.836 &0.140\\
        Multi-layer IBR-InsertNeRF & 26.89&0.915 &0.074 &25.31 &0.845 &0.131\\
        InsertNeRF (OUR)& \textbf{27.57}& \textbf{0.936}& \textbf{0.056}& \textbf{25.68}& \textbf{0.861}&\textbf{0.126} \\
        \bottomrule[2pt]
    \end{tabular}}
    \label{tab7}
    \vspace{-3mm}
\end{table}
\section{Discussion}
\subsection{Different from IBRNet about Dynamic-Static}\label{D1}

For \textit{inputs}, we employ the global-dense features as our multi-layer inputs, compared to IBRNet's single-layer dense feature, it not only retains the detailed information from dense features but also utilizes global features to predict occluded regions, as demonstrated on the depth renderings shown in~\Cref{fig1}b, where reports the rendering results of IBRNet (top) and InsertNeRF (bottom) in terms of color-depth. It is evident that InsertNeRF produces sharper edges and achieves more accurate depth predictions for background regions, even when occluded in the reference views. 

\begin{figure}[]
\begin{center}
\includegraphics[width=0.9\textwidth]{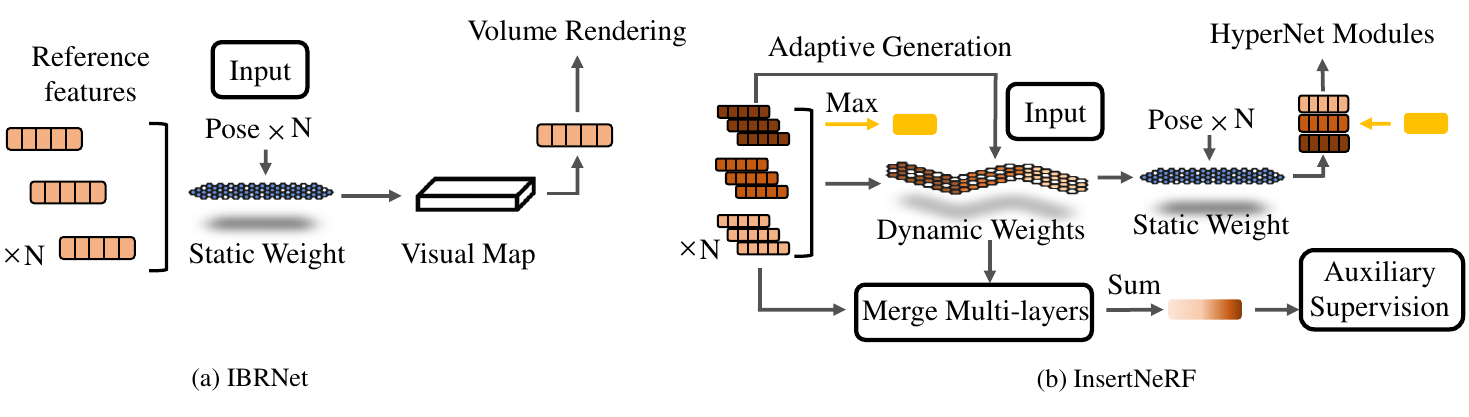} 
\vspace{-3mm}
\caption{The structures of multi-views feature aggregation parts in IBRNet and InsertNeRF}
\label{figd1}
\end{center}
\end{figure}

\begin{wrapfigure}{}{5.6cm}
\begin{center}
\includegraphics[width=0.35\textwidth]{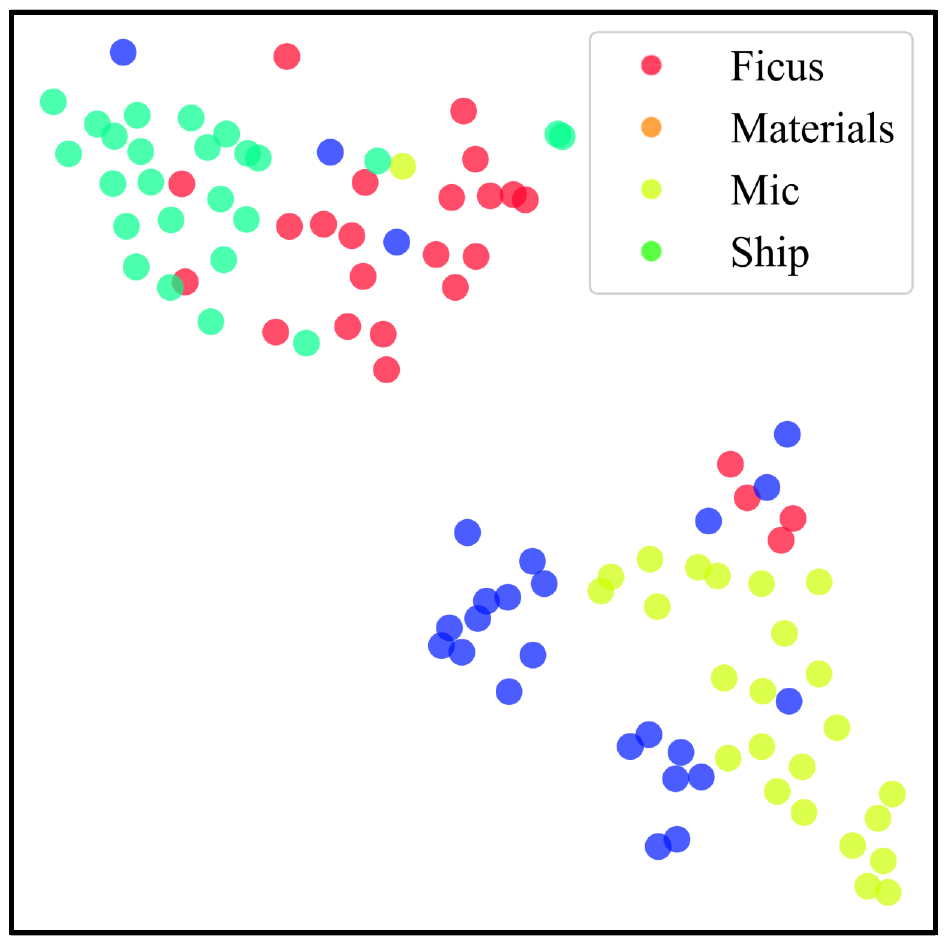} 
\caption{A t-SNE plot of the scene-specific representations in NeRF Synthetic's scenes.}
\label{fig8}
\end{center}
\end{wrapfigure}

For \textit{architecture}, IBRNet generates the visual maps using features based on the $M^{\text{ST}}$, while our InsertNeRF directly predicts $M_l^{\text{DY}}$ using multi-layer features before $M^{\text{ST}}$. Intuitively, our strategy prevents excessive reliance on the $M^{\text{ST}}$ and contributes to the adaptive inference of relationships between reference-target images. To provide additional validation, we conduct ablation experiments. Concretely, we replace the Multi-layer Dynamic-Static with the aggregation strategy from IBRNet and introduce identical multi-layer inputs into IBR-InsertNeRF for fairness. As shown in~\Cref{tab7}, despite its simplicity, our approach yields significant improvements under GNeRF settings. In addition, we also observe that the Multi-layer inputs still contributes to a notable enhancement in IBR-InsertNeRF, which is consistent with the findings in~\Cref{tab5}. 

For \textit{supervision}, in contrast to the direct generation of visual maps without any supervision, we introduce auxiliary supervision to guide $M_l^{\text{DY}}$ in fully encoding global-dense features. As shown in~\Cref{tab5}, the significance of our auxiliary supervision cannot be disregarded. In summary, as shown in~\Cref{figd1}, compared to IBRNet, the Multi-layer Dynamic-Static Aggregation Strategy focuses on predicting dynamic weights and combines them with static weight based on the multi-layer inputs and the auxiliary supervision to aggregate multiple reference features.

\subsection{Scene Representation Analysis}\label{D2}
The HyperNet modules instill generalizability into NeRF by generating scene-specific weights in the original framework. To verify this, we visualize the intermediate representations of InsertNeRF through a t-SNE plot. As shown in~\Cref{fig5}(b), it is noteworthy that the reduced-dimensional features exhibit scene clustering characteristics in LLFF evaluation data, which may be attributed to the dynamic MLPs and activation functions in our HyperNet modules. NeRF Synthetic also exhibits a similar trend. In~\Cref{fig8}, although the scenes still possess clustering characteristics, they appear relatively more dispersed, which may be attributed to significant disparities between evaluation viewpoints in the NeRF Synthetic.

\begin{table*}
    \centering
        \caption{{Ablation studies for rendering resolutions.} 
        }
    \scalebox{1}{
    \setlength{\tabcolsep}{1mm}{
    \begin{tabular}{l|ccc|ccc}
        \toprule[2pt]
            \multirow{2}{*}{Methods} & \multicolumn{3}{c|}{LLFF} & \multicolumn{3}{c}{DTU}\\
          & PSNR$\uparrow$ & SSIM$\uparrow$ & LPIPS$\downarrow$& PSNR$\uparrow$ & SSIM$\uparrow$ & LPIPS$\downarrow$\\
        \midrule
        InsertNeRF(Original Res)& 26.22& 0.838 &0.184 &29.88 &0.929 &0.096 \\
        InsertNeRF(1/2 Res) & 26.44& 0.844& 0.169  & 29.75 &0.925 &0.077 \\
        \bottomrule[2pt]
    \end{tabular}}}
    \label{tab8}
\end{table*}

\subsection{Why InsertNeRF designs could help NeRF generalization}\label{D3}

Vanilla NeRF can be considered as an implicit representation used to depict a scene through the parameters of a neural network, i.e. $\bm \Theta$, as described in~\Cref{eq1}

A natural idea is how to alter $\bm \Theta$ for different scenes $\bm{s}$, so that ${\bm \Theta}_{\bm{s}}$ possesses the ability to represent this new scene. By sampling different $\bm{s} \in \mathcal{S}$ and generating different ${\bm \Theta}_{\bm{s}}$, this can be considered as endowing vanilla NeRF representation with generalizability in multi-scenes. However, unlike explicit 3D representations such as voxels, meshes, and point clouds, constructing an implicit representation ${\bm \Theta}_{\bm{s}}$ directly for a given $\bm{s}$ is challenging. 

Therefore, in this paper, we introduce the HyperNet modules, which is invented to generate weights for a target neural network, to address this issue. Through two types of the HyperNet modules we propose, scene-customization weights (parameters) $\bm\Omega_T^{\bm{s}}$ in the NeRF framework are generated in a given $\bm{s}$. Here, we predict $\bm\Omega_T^{\bm{s}}$ by combining the feature extraction from reference images and the  multi-Layer dynamic-static aggregation strategy. Finally, by combining $\bm \Theta$ and new weights $\bm\Omega_T^{\bm{s}}$ within the NeRF framework, we obtain ${\bm \Theta}_{\bm{s}}$ that can adapt to different scenes $\bm{s}$, as described in~\Cref{InsertNeRF}

\section{Additional Results and Analysis}\label{E}
\subsection{Additional Ablation Studies}
\textbf{Rendering Resolutions}: During the evaluation phase, prior works employed different rendering resolutions, which has some impact on the metric. To investigate this issue, we evaluate the rendering performance at different resolutions without altering the training settings. In~\Cref{tab8}, reducing the rendering resolution not only improved rendering efficiency but also demonstrated performance enhancements in LLFF. However, in the DTU dataset, a contrasting trend is evident, which may be attributed to its emphasis on foreground rendering. 

\begin{wraptable}{r}{0.4\textwidth}
    \centering
    \vspace{-5mm}
    \caption{{Ablation studies for $\mathcal{F}_{app}$.}}
    \vspace{-2mm}
    \scalebox{0.85}{
    \begin{tabular}{c|ccc}
        \toprule[2pt]
        \multirow{2}{*}{$Z$} & \multicolumn{3}{c}{DTU}\\
          & PSNR$\uparrow$ & SSIM$\uparrow$ & LPIPS$\downarrow$\\
        \midrule
        $Z=1$ ($\mathcal{F}_{geo}$)  &29.03& 0.918& 0.086 \\
        $Z=2$ &29.75& 0.925& 0.077\\
        $Z=3$ &29.83& 0.925& 0.075\\
        \bottomrule[2pt]
    \end{tabular}}
    \label{tab13}
\end{wraptable}

\textbf{Parallel Branches $Z$}: We further analyze the influence of the number of parallel branches $Z$ on network rendering performance, as shown in~\Cref{tab13}. When $Z=1$, i.e., $\mathcal{F}_{app}$ is replaced by $\mathcal{F}_{geo}$ with the same input, a significant performance drop occurs. This might be attributed to the implicit modeling of BRDF prior by $\mathcal{F}_{app}$. Furthermore, with an increase in the number of branches, $\mathcal{F}_{app}$ endows the NeRF framework with enhanced fine-detail generalizability. 

\begin{table*}
    \centering
        \caption{{Comparisons of HyperNet modules against SOTA methods on ShapeNet.} 
        }
    \scalebox{0.95}{
    \setlength{\tabcolsep}{1mm}{
    \begin{tabular}{l|cc|cc|cc|cc}
        \toprule[2pt]
            \multirow{2}{*}{Methods} & \multicolumn{2}{c|}{Chairs 1-view}& \multicolumn{2}{c|}{Chairs 2-views}  & \multicolumn{2}{c|}{Cars 1-view}& \multicolumn{2}{c}{Cars 2-views}\\
          & PSNR$\uparrow$ & SSIM$\uparrow$ & PSNR$\uparrow$ & SSIM$\uparrow$ & PSNR$\uparrow$ & SSIM$\uparrow$ & PSNR$\uparrow$ & SSIM$\uparrow$\\
        \midrule
        ENR~\citep{dupont2020equivariant}& 22.83& - &- &- &22.26 & - &- &- \\
        SRN~\citep{sitzmann2019scene}& 22.89& 0.89 &24.48 &0.92 &22.25 & 0.89 &24.84 &0.92 \\
        ViewFormer (ECCV 2022)& 14.74& 0.79 &17.20 &0.84 &19.03 & 0.83 &20.09 &0.85 \\
        pixelNeRF (CVPR 2021)& 23.72& 0.91 &26.20&0.94 &23.17 & 0.90 &25.66 &0.94 \\
        pixelNeRF+HyperNet modules & \textbf{24.51}& \textbf{0.92} &\textbf{26.71}&\textbf{0.94} &\textbf{24.18} & \textbf{0.91} &\textbf{26.05} &\textbf{0.95} \\
        \bottomrule[2pt]
    \end{tabular}}}
    \label{tabShapeNet}
\end{table*}

\subsection{Results for ShapeNet}
In this section, we explore the performance of the our InsertNeRF in ShapeNet under Chairs and Cars scenes. Due to the primary emphasis of InsertNeRF on multi-view settings I and II, the validation for multi-layer dynamic-static aggregation strategy under few-views settings is unnecessary. Therefore, we integrate the HyperNet modules into the original pixelNeRF~\cite{yu2021pixelnerf}, altering its training inputs accordingly. As shown in the~\Cref{tabShapeNet}, our modules exhibit significant improvements compared to pixelNeRF~\cite{yu2021pixelnerf}, especially in the 1-view setting. It's also evident that compared to NeRF Synthetic, LLFF, and DTU, InsertNeRF shows less improvement on ShapeNet. This might be due to the relatively simplistic appearance and geometry of ShapeNet-Scenes, and our work primarily focuses on multi-view settings as mentioned in~\citep{kulhanek2022viewformer}.

\subsection{Results from Fine-Tuning}
We also explore the rendering performance of InsertNeRF after fine-tuning in various scenes. In contrast to the fine-tuning methodology adopted by~\citep{liu2022neural}, we fine-tune directly on the pre-trained model. \Cref{tab9} presents the performance of InsertNeRF across different scenes and the results after fine-tuning in NeRF Synthetic.

\begin{table*}
    \caption{The performance in different scenes and the results after Fine-Tuning in NeRF Synthetic.}
    \centering{
    \subtable[PSNR]{
    \scalebox{0.95}{
    \begin{tabular}{l|l|ccccccccc}
        \toprule[2pt]
        Method &FT & Lego & Chair & Drums &  Ficus & Hotdog & Materials &Mic &Ship &Avg.\\
        \midrule
          InsertNeRF & &30.00    &  31.97 &  25.59&  26.08&  35.04 &  28.91 &  35.13 &  30.08 & 30.35\\
           MVSNeRF &\checkmark   & -    &  - &  -&  -&  - &  - &  - &  -& 27.21\\
           IBRNet&\checkmark& -    &  - &  -&  -&  - &  - &  - &  -& 30.05\\
           GNT&\checkmark& 31.38 & 33.70 &26.98&29.55&36.95 &29.11 &33.35 &30.54& 31.45\\
          NeuRay &\checkmark & -    &  - &  -&  -&  - &  - &  - &  -& 32.35\\
         InsertNeRF&\checkmark  & 31.93  & 34.41  & 27.76 &29.08 &37.25 &31.46 & 36.43 &31.98 &\textbf{32.54}\\
        \bottomrule[2pt]
    \end{tabular}}}}

    \centering{
    \subtable[SSIM]{
    \scalebox{0.95}{
    \begin{tabular}{l|l|ccccccccc}
        \toprule[2pt]
        Method &FT & Lego & Chair & Drums &  Ficus & Hotdog & Materials &Mic &Ship &Avg.\\
        \midrule
          InsertNeRF & &0.939    & 0.969 & 0.910&  0.915&  0.969 &  0.922 &  0.972 &  0.888 & 0.936\\
           MVSNeRF &\checkmark   & -    &  - &  -&  -&  - &  - &  - &  -& 0.888\\
           IBRNet&\checkmark& -    &  - &  -&  -&  - &  - &  - &  -& 0.935\\
          NeuRay &\checkmark & -    &  - &  -&  -&  - &  - &  - &  -& 0.960\\
         InsertNeRF&\checkmark  & 0.962  & 0.981  & 0.937 &0.959 &0.988 &0.957 & 0.978 &0.931 &\textbf{0.962}\\
        \bottomrule[2pt]
    \end{tabular}}}}
    \centering{
    \subtable[LPIPS]{
    \scalebox{0.95}{
    \begin{tabular}{l|l|ccccccccc}
        \toprule[2pt]
        Method &FT & Lego & Chair & Drums &  Ficus & Hotdog & Materials &Mic &Ship &Avg.\\
        \midrule
          InsertNeRF & &0.056    &  0.033 &  0.079&  0.085&  0.037 &  0.076 &  0.029 &  0.129 & 0.066\\
           MVSNeRF &\checkmark   & -    &  - &  -&  -&  - &  - &  - &  -& 0.162\\
           IBRNet&\checkmark& -    &  - &  -&  -&  - &  - &  - &  -& 0.066\\
          NeuRay &\checkmark & -    &  - &  -&  -&  - &  - &  - &  -& 0.048\\
         InsertNeRF&\checkmark  & 0.044  & 0.024  & 0.065 &0.054 &0.027 &0.048 & 0.021 &0.102 &\textbf{0.048}\\
        \bottomrule[2pt]
    \end{tabular}}}}
    \label{tab9}
\end{table*}

\begin{table*}
    \caption{Comparison of InsertNeRF for single scene rendering on the NeRF Synthetic.}
    \centering{
    \subtable[PSNR]{
    \scalebox{1}{
    \begin{tabular}{l|cccccccc}
        \toprule[2pt]
        Method  & Room & Fern & Leaves &  Fortress & Orchids & Flower &T-Rex &Horns \\
        \midrule
          LLFF  &24.54    & \textbf{28.72} & 21.13& 21.79& 18.52 &  20.72 &  27.48 &  23.22 \\
           NeRF    & \underline{32.70}    &\underline{25.17} & 20.92& 31.16&20.36 & \underline{27.40} &26.80 & 27.45\\
          GNT  & \textbf{32.96} &  24.31  & \underline{22.57}& \textbf{32.28}& \underline{20.67} &  27.32& \textbf{28.15} & \textbf{29.62} \\
         InsertNeRF  & 32.55  & 24.88  & \textbf{22.59} &\underline{31.82} &\textbf{21.18} &\textbf{28.39} & \underline{27.49} &\underline{29.37} \\
        \bottomrule[2pt]
    \end{tabular}}}}
    \centering{
    \subtable[SSIM]{
    \scalebox{1}{
    \begin{tabular}{l|cccccccc}
        \toprule[2pt]
        Method  & Room & Fern & Leaves &  Fortress & Orchids & Flower &T-Rex &Horns \\
        \midrule
          LLFF  &0.932    & 0.753 & 0.697& 0.872& 0.588 & 0.844 & 0.857 & 0.840 \\
           NeRF    & 0.948    &0.792 & 0.690& 0.881&0.641 & 0.827 &0.880 & 0.828\\
          GNT  & \textbf{0.963} &  \textbf{0.846}  & \underline{0.852}& \textbf{0.934}& \underline{0.752} & \underline{0.893}& \textbf{0.936} & \textbf{0.935}\\
         InsertNeRF  & \underline{0.961}  & \textbf{0.846}  & \textbf{0.853} &\underline{0.925} &\textbf{0.756} &\textbf{0.904} & \underline{0.928} &\underline{0.932} \\
        \bottomrule[2pt]
    \end{tabular}}}}
    \centering{
    \subtable[SSIM]{
    \scalebox{1}{
    \begin{tabular}{l|cccccccc}
        \toprule[2pt]
        Method  & Room & Fern & Leaves &  Fortress & Orchids & Flower &T-Rex &Horns \\
        \midrule
          LLFF  &0.155    & 0.247 & 0.216& 0.173& 0.313 & 0.174 & 0.222 & 0.193 \\
           NeRF    & 0.178    &0.280 & 0.316& 0.171&0.321 & 0.219 &0.249 & 0.268\\
          GNT  & \textbf{0.060} &  \textbf{0.116}  & \textbf{0.109}& \textbf{0.061}& \underline{0.153} & \underline{0.092}& \textbf{0.080} & \underline{0.076}\\
         InsertNeRF  & \underline{0.063}  & \underline{0.121}  & \textbf{0.109} &\underline{0.062} &\textbf{0.152} &\textbf{0.070} & \underline{0.085} &\textbf{0.074} \\
        \bottomrule[2pt]
    \end{tabular}}}}
    \label{tab10}
\end{table*}


\subsection{Single Scene Results}
Existing works~\citep{wang2022attention} also tend to focus on single-scene rendering within the framework of GNeRF. We conduct a quantitative comparison with existing works in the single-scene setting and achieve satisfactory performance, as shown in~\Cref{tab10}.

\subsection{More Qualitative Results}\label{E4}
\begin{figure}[t]
\begin{center}{
\includegraphics[width=1\textwidth]{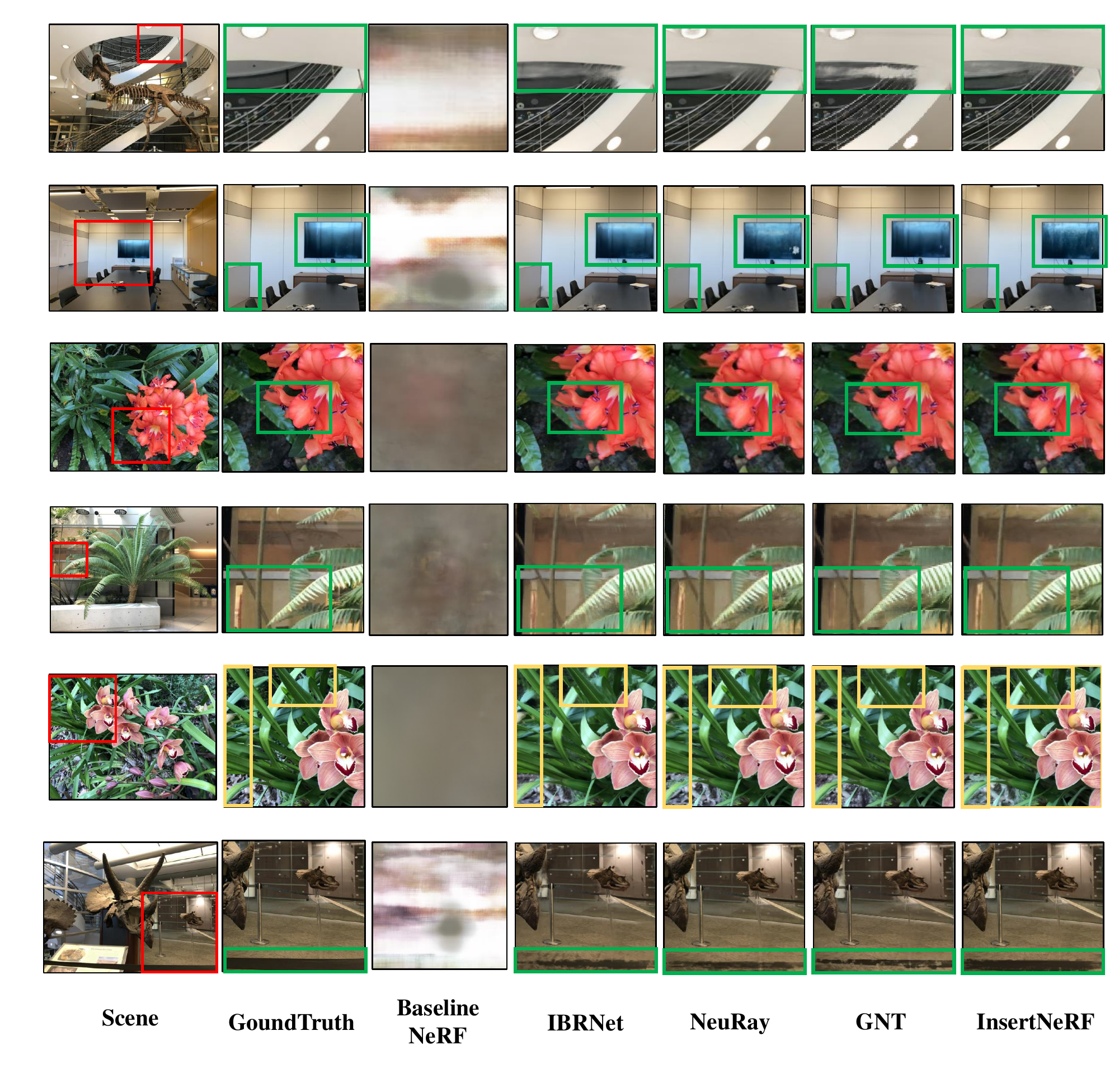} 
\caption{Qualitative comparisons of InsertNeRF against SOTA methods under LLFF scenes.} 
\label{fig9}}
\end{center}
\end{figure}

\begin{figure}[t]
\begin{center}{
\includegraphics[width=1\textwidth]{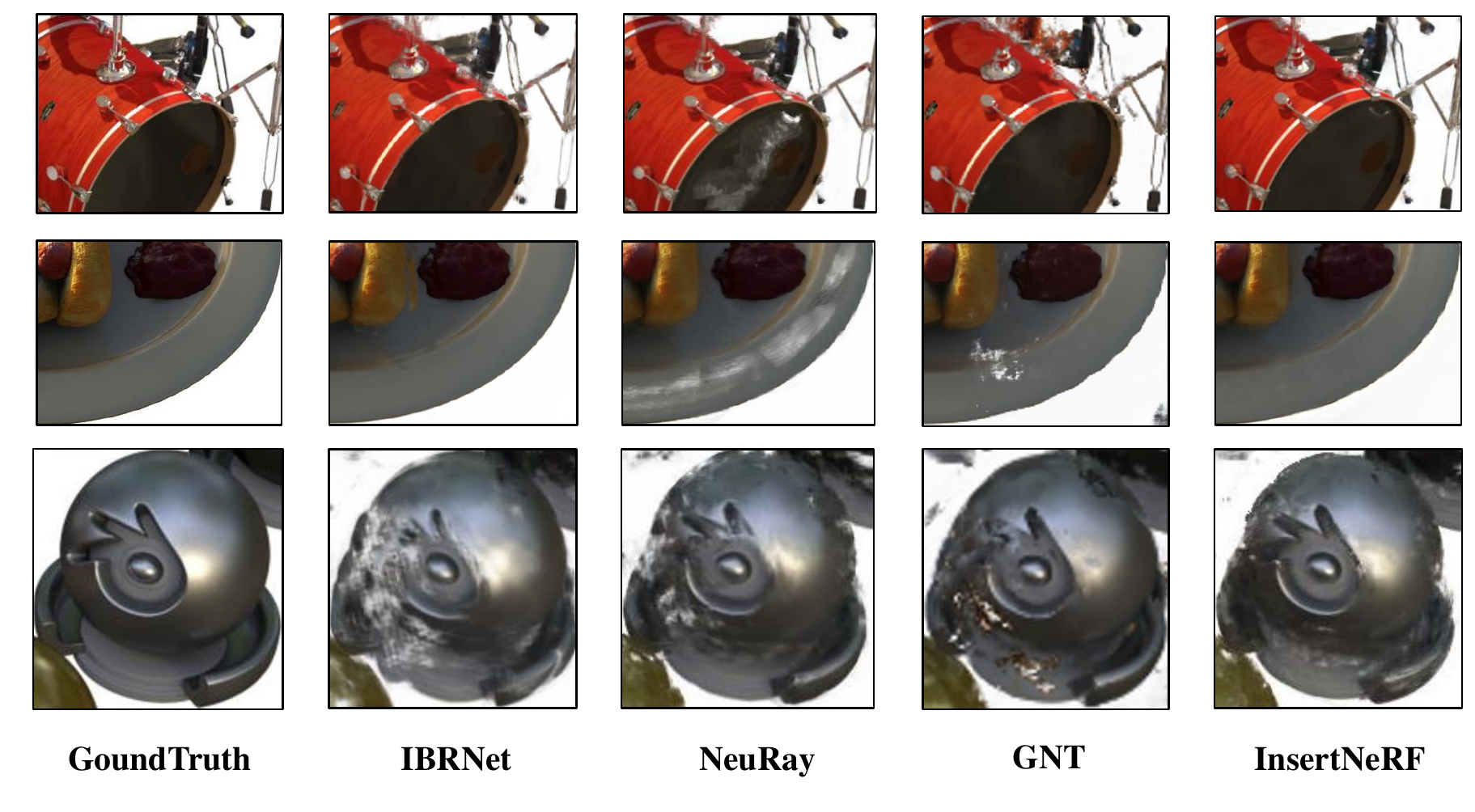} 
\caption{Qualitative comparisons of InsertNeRF against SOTA methods under NeRF Synthetic scenes.} 
\label{fig10}}
\end{center}
\end{figure}

\begin{figure}[t]
\begin{center}{
\includegraphics[width=1\textwidth]{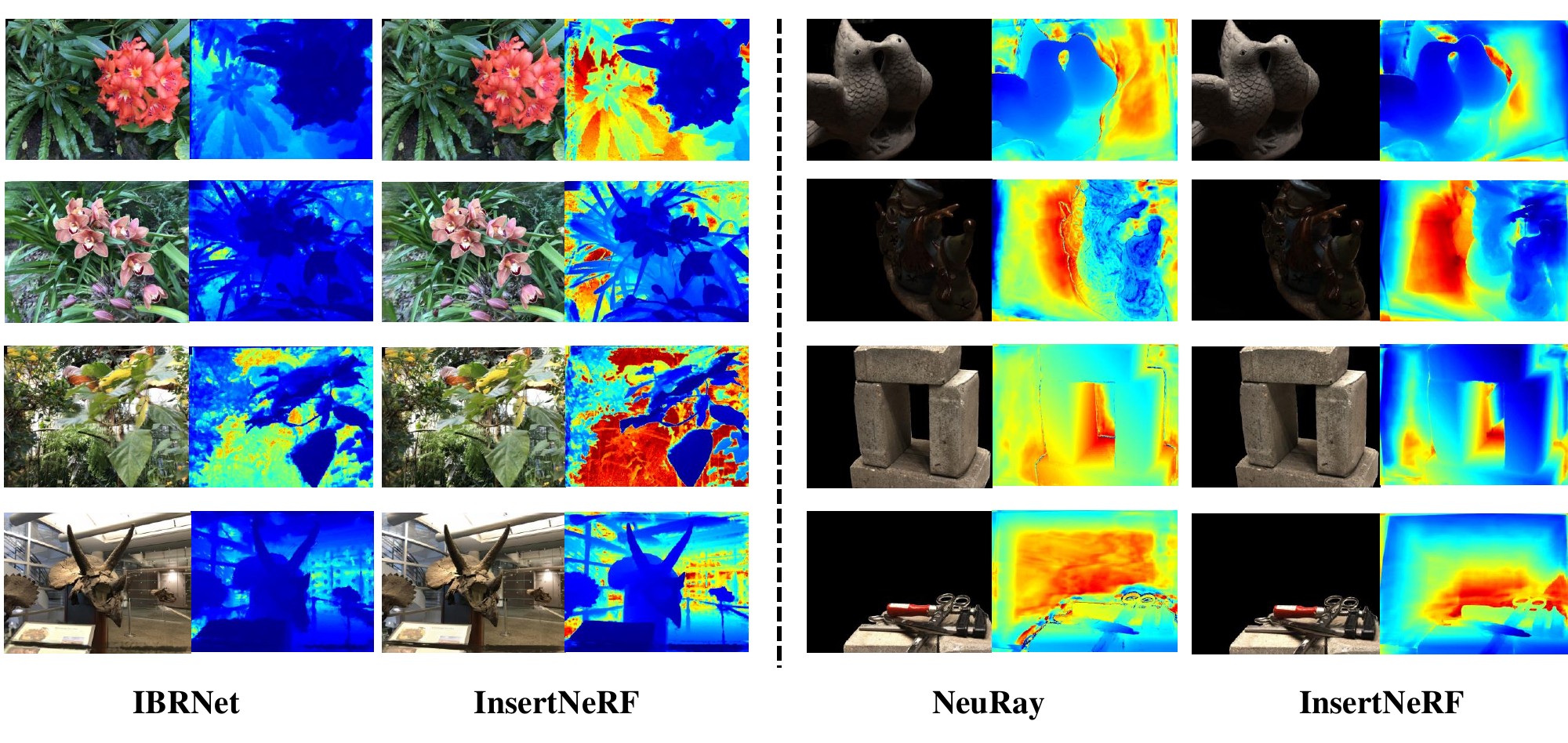} 

\caption{Color-Depth results of InsertNeRF with existing works under LLFF and DTU's scenes.} 
\label{fig11}}
\end{center}
\end{figure}

\begin{figure}[t]
\begin{center}{
\includegraphics[width=0.9\textwidth]{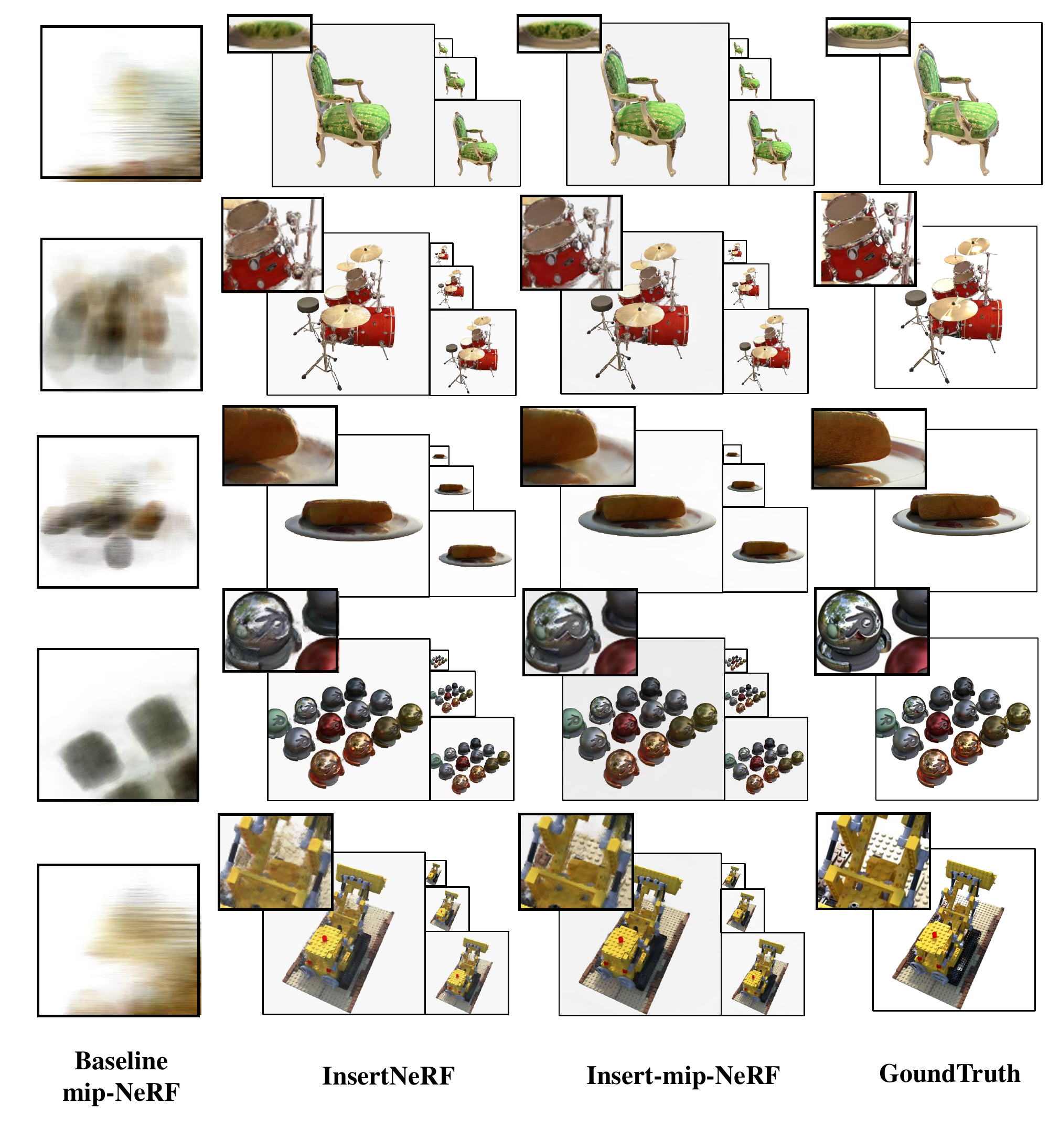} 
\caption{Qualitative results of Insert-mip-NeRF in multi-scale NeRF Synthetic.} 
\label{fig12}}
\end{center}
\end{figure}

\begin{figure}[t]
\begin{center}{
\includegraphics[width=0.85\textwidth]{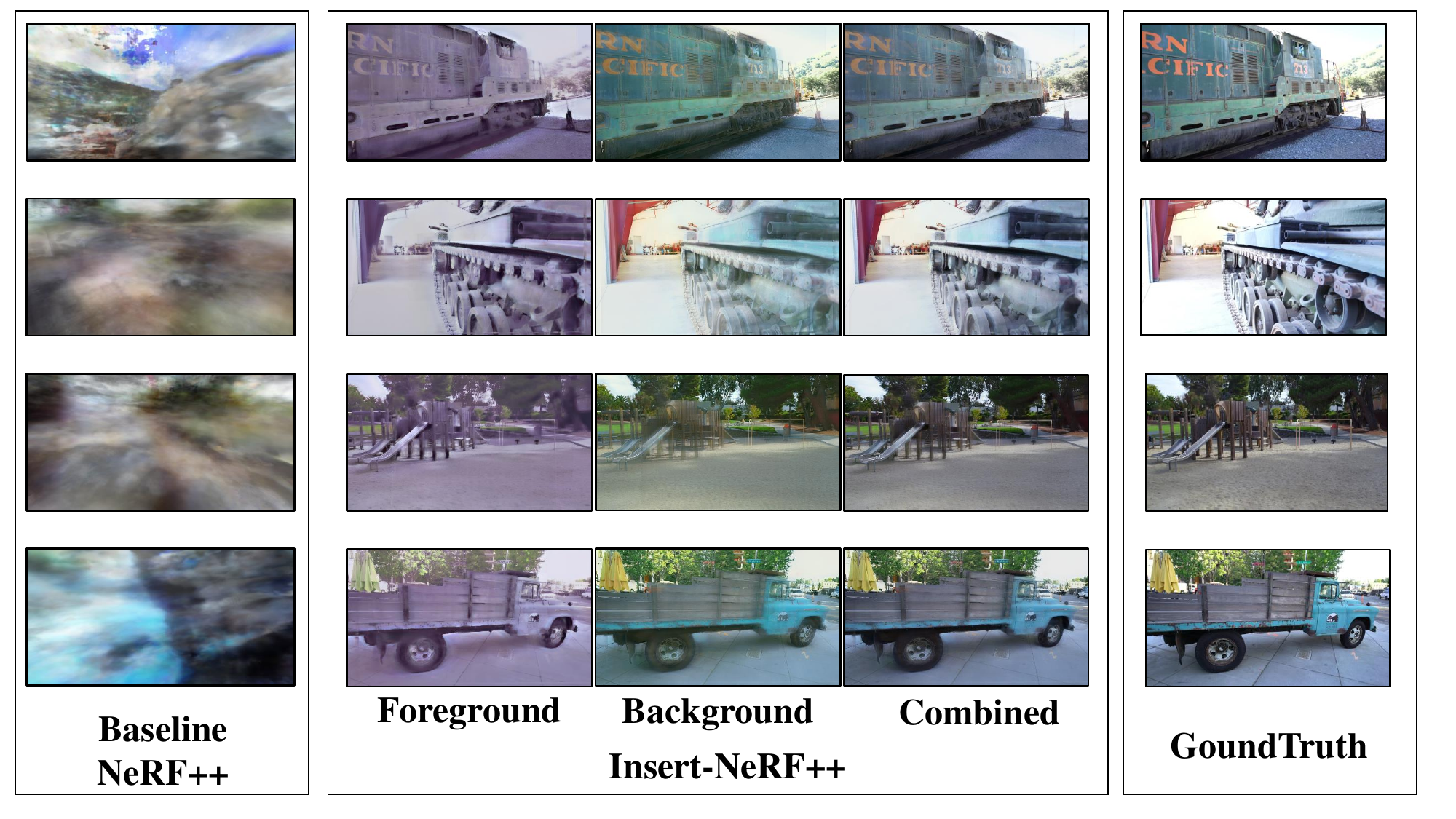} 

\caption{Qualitative results of Insert-NeRF++ in Tanks and Temples. } 
\label{fig13}}
\end{center}
\end{figure}

We present additional qualitative results to further analyze the superiority of InsertNeRF. i). \Cref{fig9} and~\Cref{fig10} report qualitative results in LLFF and NeRF Synthetic. ii). ~\Cref{fig11} showcase more Color-Depth results in LLFF and DTU datasets. iii). We also qualitatively analyze the generalizability of the InsertNeRF-systems including Insert-mip-NeRF~\Cref{fig12}, and Insert-NeRF++~\Cref{fig13}. Note that the presence of color distortion in the Insert-NeRF++'s foreground rendering is observed, yet it does not impact the combined results, possibly attributable to the replaced sampling process.

\section{Limitations}
In the majority of scenarios, a higher number of sample points along the rays often leads to improved rendering performance. In essence, thanks to the transformer architecture, existing works~\citep{wang2022attention, wang2021ibrnet} can be trained on a limited number of sample points (64 training samples) and evaluated on the more sample points (192 evaluation samples), resulting in elevated training efficiency and improved rendering performance. However, in the InsertNeRF-system, it is essential to maintain consistency in the number of sample points between the training and evaluation processes. In order to ensure fairness in comparative experiments and strike the trade-off between training efficiency and rendering performance, we set $K=64$ both during training and evaluation, which imposes certain limitations on the rendering performance. Naturally, as we increase the number of sample points for training, the rendering performance will further improve.

\section{Future Work}
We aspire to construct an all-encompassing InsertNeRF framework, endowing generalizability into various NeRF-derived works, such as TensoRF, NeRF--, NeuS, and so forth. This can facilitate existing or future NeRF research to transcend the constraints of scene-customization.

In addition, we have provided a pre-trained model to address NeRF under sparse inputs in~\Cref{sec:more-frameworks}. While such models have demonstrated satisfactory performance without any retraining~\Cref{tab3}, we still plan to design a fine-tuning approach for sparse inputs to further enhance rendering quality.

\end{document}